%% file: main.tex
\documentclass[12pt, letter]{article}

\usepackage[T1]{fontenc}
\usepackage{lmodern}

\usepackage{tikz}
\usepackage{siunitx}
\usepackage{dirtytalk}
\usepackage{quoting}
\quotingsetup{vskip=5pt}
\usepackage{comment}
\usepackage{algorithm, algcompatible}
\usepackage[noend]{algpseudocode}
\usepackage{csvsimple}
\usepackage{listings}
\usepackage{biblatex}
\usepackage{authblk}
\usepackage[letterpaper, portrait, margin=0.75in]{geometry}
\usepackage{graphicx}
\usepackage{pdflscape}
\usepackage[symbol]{footmisc}
\usepackage{fixfoot}
\usepackage{subcaption}
\usepackage{mwe}
\usepackage{csquotes}
\usepackage{hyperref}

\makeatletter
\patchcmd{\@maketitle}{\LARGE}{\Large}{}{}
\makeatother

\renewcommand{\thefootnote}{\fnsymbol{footnote}}

\title{Transforming Unstructured Text into Data with Context Rule Assisted Machine Learning (CRAML)}

\renewcommand{\thefootnote}{\fnsymbol{footnote}}

\author{Stephen Meisenbacher\protect\footnotemark \protect\footnotetext{Equally contributing authors. We are grateful for support from the Economic Security Project Anti-Monopoly Fund; Loyola Rule of Law Institute; Loyola Quinlan School of Business; and Loyola University Chicago. We also thank Patricia Tabarani, Eric George, Steve Sauerwald, and Chris Erickson.} \footnote{Technical University of Munich, School of Computation, Information and Technology} \and Peter Norlander\protect\footnotemark[1] \footnote{Loyola University Chicago, Quinlan School of Business}} 

\begin{document}

\maketitle 

\renewcommand{\thefootnote}{\arabic{footnote}}

\begin{abstract}

\textbf{Abstract}: We describe a method and new no-code software tools enabling domain experts to build custom structured, labeled datasets from the unstructured text of documents and build niche machine learning text classification models traceable to expert-written rules. The Context Rule Assisted Machine Learning (CRAML) method allows accurate and reproducible labeling of massive volumes of unstructured text. CRAML enables domain experts to access uncommon constructs buried within a document corpus, and avoids limitations of current computational approaches that often lack context, transparency, and interpetability. In this research methods paper, we present three use cases for CRAML: we analyze recent management literature that draws from text data, describe and release new machine learning models from an analysis of proprietary job advertisement text, and present findings of social and economic interest from a public corpus of franchise documents. CRAML produces document-level coded tabular datasets that can be used for quantitative academic research, and allows qualitative researchers to scale niche classification schemes over massive text data. CRAML is a low-resource, flexible, and scalable  methodology for building training data for supervised ML. We make available as open-source resources: the software, job advertisement text classifiers, a novel corpus of franchise documents, and a fully replicable start-to-finish trained example in the context of no poach clauses. 
\end{abstract}

\clearpage

\input{intro.tex}

\input{challenges.tex}

\input{motivation.tex}

\input{data_analysis.tex}

\input{discussion.tex}

\newpage

\section*{Acknowledgements}
We are grateful for support from the Economic Security Project Anti-Monopoly Fund; Loyola Rule of Law Institute; Loyola Quinlan School of Business; and Loyola University Chicago. All errors are the authors'.

\section*{Conflict of interest}
The authors have no conflicts of interest.

\section*{Supporting Information}

See our Github repository for the full code base: \url{https://github.com/sjmeis/CRAML_Beta}.

\clearpage
\nocite{*}
\printbibliography

\appendix
\clearpage
\input{appendixa.tex}

\clearpage
\input{appendixb.tex}

\clearpage
\input{appendixc.tex}

\end{document}

%% file: intro.tex
\section{Introduction}
Advances in computational methods offer new ways to gain insight from large volumes of unstructured text, and yet these methods each have significant limitations, tradeoffs, and a lack clear guidelines \cite{pandey_applying_2019}. Despite the advances in Natural Language Processing (NLP), Machine Learning (ML), Artificial Intelligence (AI) and more recently Deep Learning (DL), there is still ``the herculean task of finding constructs using full-text search'' \cite[547]{larsen_tool_2016}.

%extracting meaningful information from massive amounts of unstructured text data. 

Unstructured text lack a schema, are not standardized, have multiple formats, and come from diverse sources \cite{adnan_limitations_2019}. Greater attention and systematic research is needed to develop processes to create structured data from unstructured text \cite{dimaggio_adapting_2015}. A lack of structured information is a barrier to understanding for researchers and organizations: ``as much as 80\% of an organization’s data is `dark''' \cite{lacity_becoming_2021}. For knowledge professionals, just-in-time access to domain specific document repositories is vital, but knowledge must first be codified \cite{subramani_capability_2021}. Zettabytes (ZB) of new data are produced daily \cite{8399125}, and roughly 80\% is unstructured \cite{hammoud2019personal}. While the manual expert labor required for qualitatively coding novel classification schemes is hard to scale, statistical packages, many information systems, and quantitative social scientists expect data to be codified and in a regular format: “tabular data – variables in columns, cases in rows” \cite{lazer_data_2017}. 

ML and AI are potential solutions, but require structured training data. Hand-coding by experts is required in many domains, but does not easily scale. To address the disjunction between these, we describe a method that bridges expert-built set of context rules scaled to classify unstructured text and build training data for machine learning. Context Rule Assisted Machine Learning (CRAML) is a hybrid method for structuring textual data from large-scale corpora into datasets ready for training ML models and quantitative research. CRAML begins with a workflow common in qualitative manual coding research, and ends with ML models trained on text input data shaped by a qualitative coding scheme. CRAML is useful for qualitative and quantitative social scientists: it empowers domain experts to efficiently mine volumes of text and scale classification schemata, and it yields tabular data that captures the occurrence of concepts within a corpus of documents. 

This paper contributes to information systems research on the design of intelligent systems that can generate economic and social value from unstructured data \cite{abbas_text_2018}. By clearly defining the problem, managing the training data before AI is attempted, and affording users the opportunity to evaluate and adjust both inputs and outputs, CRAML addresses major concerns identified in research on AI \cite{zhang_addressing_2020}. This is the first paper to offer a thorough description of a versatile method with the potential to empower advanced users of an ordinary desktop computer to build machine learning classifiers and rectangular datasets from unstructured text according to the users' desired scheme. Expert built rule-assisted machine learning models have been described in the literature, and found to outperform benchmark methods and professional perception in critical cases such as detecting persons in states of emotional distress \cite{chau_finding_2020}. In addition to describing the method, we call and set new standards for transparency and interpretability of research on text-to-data machine learning, and demonstrate the high utility of this approach in three applications. 

%and often relies on standardized benchmark datasets, proprietary data, and methods that fail to reproduce reliable results \cite{kapoor_narayanan}

%We contribute and document a novel process and new software tools by which an individual domain expert or a small team can extract, sample, analyze, and hand-code contextual rules, and validate and extrapolate the rules to build structured datasets and ML classifiers. CRAML's pipeline provides qualitative and quantitative researchers new no-code software tools and a step-by-step method to sift through massive volumes of text within a corpus\footnote{Documents that belong to a collection, or corpus, of similar documents are often referred to as \textit{unstructured}, in the sense that no inherent structured information is present. For example, tweets are \say{documents} continuously being created and added to the text corpus of Twitter. Likewise, complex legal contracts and job advertisements are documents that are regularly amassed into large, domain-specific corpora.}, analyze and learn patterns, and mold a structured dataset by building context rules \textit{in the worldview} of the expert.

In this methods paper, we highlight the need for the CRAML tool and demonstrate its versatility through applications. First, we use CRAML to perform a simple literature review that highlights current approaches to analysis of large text corpora in five top management journals. Second, we release machine learning classifiers with the potential to enhance labor market efficiency, and set a new standard for transparency for research studying job advertisement text, where underlying data includes proprietary data obtained under license. We release ML classifiers, and the rules that created them, to enable other researchers using job advertisement text to replicate and interpret our approach. Last, we publish a new and massive document corpus of mandatory franchise disclosure documents and present a start-to-finish analysis to enable a full replication of our process. We detect thousands more plausibly illegal ``no poach clauses'' that could violate state and federal antitrust law than found in earlier research \cite{krueger_theory_2022}. We release a rectangular dataset that represents the prevalence of suspected no poach clauses that restrict the ability of franchise firms to recruit employees from other companies that belong to a franchise system.  As part of the contribution of this work, we publish everything needed to fully replicate the no poach analysis: the original PDFs and the machine readable text of the franchise documents, CRAML software, the manually coded inputs that creates the structured, labeled data output, the training data, and a ML classifier to detect suspect no poach clauses.

%We release the documents in machine-readable format, the CRAML software, and all components of the analysis to enable replication. 

Section \ref{sec:sota} reviews the well-known challenges CRAML was created to address and illustrates the frequency of current approaches in the management literature. Section \ref{ref:motivate} introduces and provides detailed motivation for our solution in the context of job advertisement text. Methodological detail on CRAML and empirical analysis of franchise no poach agreements is in Section \ref{sec:analysis}. A more technical methodology is described in Appendix \ref{sec:appendixc}. Section \ref{sec:discuss} provides future directions and discussions.

%In the \say{Modern ML Paradigm,} the process for structuring text data for ML before analysis often relies on standardized benchmark datasets or proprietary data and methods to interpret a fire hose of text, which comes at the expense of \emph{interpretability} and \emph{replicability} and limits the ability to apply ML to real-world challenges \cite{nelson_measure_2019}. 

%% file: challenges.tex
\section{The State of the Art}
\label{sec:sota}
Across many fields of research, methods for analyzing unstructured text are needed. Below, we highlight literature in management, information systems, computational social science and computer science on: machine learning (ML), artificial intelligence (AI), natural language processing (NLP), topic modeling and Latent Dirichilet Allocation (LDA), linguistic inquiry and word counting (LIWC), computationally aided text analysis (CATA), and text or document corpora. We used the CRAML software to narrow our search for relevant literature. We obtained a corpus of 1,994 papers published from 2015-2021 in 5 top empirical management journals (Academy of Management Journal, Administrative Science Quarterly,  Journal of International Business Studies, Organization Science, and Strategic Management Journal). Within the management literature corpus,  we searched for papers that could illustrate the use of a method for turning unstructured text into data. In Appendix \ref{sec:appendixa}, we provide additional detail on how CRAML can be used in bibliographic research. 

\subsection{Word Counting Approaches (LIWC, CATA)}
\label{sec:sifting}

Computer-aided text analysis (CATA) and Linguistic Inquiry and Word Count (LIWC) are two popular techniques used in the management literature. Experts develop lists of words or dictionaries that correspond to an overall construct and measure the occurrence of keywords inside specific documents \cite{short_construct_2010,short_application_2008}. Researchers often use LIWC standard dictionaries to detect common constructs, such as positive / negative sentiment in a corpus of tweets \cite{bachura_opm_2022}. Using these methods, researchers can also develop custom dictionaries to capture previously unstudied phenomena within an environment: for example, to measure how actors use specific cultural toolkits in a field of action  \cite{weber_toolkit_2005}. Such efforts are typically first driven by deduction (keyword selection) and then interrogation of text (examining n-grams and short excerpts of text containing keywords) \cite{mckenny_using_2013}. 

\begin{figure}[ht]
\centering
\includegraphics[scale=0.6]{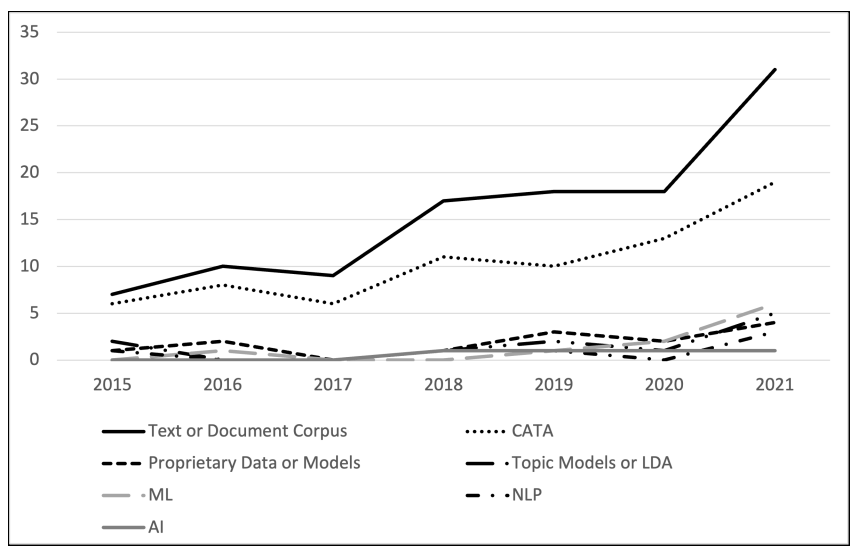}
\caption{Counting Words: The Growth of Management Papers Using Text Data and CATA, AI, ML, LDA, NLP}  
\label{fig:frequency_of_words}
\end{figure}

We developed keywords deductively (available in Appendix \ref{sec:appendixa}) and narrowed our search for relevant papers using CRAML. CRAML output indicates whether each academic paper in the corpus contains keywords that fall within a given construct (or tag): we focused on a subset of 110 papers that contain reference to text data or a document corpus.  Figure \ref{fig:frequency_of_words} shows that text corpus analysis and CATA along with it have grown over time; 88 of the 110 papers that mention a text corpus also refer to CATA.

\subsubsection{Limitations of Keyword-Based Approaches}

Conceptual schemes that rely only on keywords are not very rich \cite{duriau_content_2007}. Constructs may not be well-captured by word frequency approaches used in CATA and LIWC programs; these methods perform less well than ML at tasks such as inferring personality \cite{cutler_inferring_2021}. Word counting  assumes that all word occurrences equally contribute to the measurement of a construct - that every use of the keyword ``artificial intelligence'' or ``machine learning'' captures both papers that use an AI or ML method, as well as those that study or mention AI and ML phenomena. Because the method relies on exact matching and correct spellings, CATA and LIWC will fail to recognize concepts when faced with messy data or typos. Without the benefit of context surrounding the keywords, key meaning and accuracy are lost to the user. When describing the frequency of terms inside a document, CATA methods do not typically adjust for document length, although additional steps can be taken to make adjustments \cite{guo_impact_2021}.

While keywords can start a search for relevant concepts, rules and qualitative coding pinpoint only the relevant text within the documents. Qualitative researchers who seek to inductively build theory typically begin by engaging in keyword search, but also engage in qualitative hand coding \cite{li_digital_2022}. 

%We initially found 1,213 papers contain keywords potentially relevant to one of the concepts in our discussion below (proprietary data and models). However, based on n-gram analysis of chunks extracted, we develop sets of rules based on qualitative coding (also in Appendix \ref{sec:appendixa}) in CRAML to indicate which papers are relevant to the phenomena within our scope of interest. For example, with a regular expression, we narrow the proprietary data or models construct to include only the papers that include keywords related to proprietary information, and either ``data'' or ``model'' within a 13 word chunk of text extracted from the management corpus. 

\subsection{Unsupervised Learning (Topic Modeling and LDA)}

Text clustering aims to solve the problem of information overload by simplifying a mass of text into a smaller number of meaningful categories. While many text clustering methods exist \cite{aggarwal2012survey}, a popular technique is \textit{topic modeling}, which aims to learn \say{topics} from an unstructured collection of text documents. Data is often first heavily pre-processed (stemmed), stripped of common words, and converted into a document-term matrix, which leaves only the stems of latent keywords appearing. Latent Dirichlet Allocation (LDA) assigns all words in a document to a topic, under the assumption that every part of every document contributes to some topic. LDA's versatility is that it is not domain-specific: it can be used to study hacker forums \cite{yue_see_2019}, why people retweet \cite{geva_using_2019}, online communities \cite{bapna_nurturing_2019}, and discourse on emerging technologies like blockchain \cite{miranda_discursive_2022}. LDA can also assist in inductive theorizing \cite{shrestha_algorithm_2021}.

%, NLP, AI, or supervised ML. Of the 110 text corpus papers in the management literature, only 31 papers discuss the use of one or more of these methods. However, the use of these techniques has increased recently, with 15 of these papers appearing in 2021. 

Of the 110 text corpus papers in the management literature,  11 papers discuss the use of this method, with 5 of these papers appearing in 2021. \cite{choi_using_2021} use LDA to arrive at text-based measures of corporate dissimilarity from corporate annual reports, and use 25 crowd-sourced MTurkers to ensure themes are reliably identified for the top 25 of 125 latent topics chosen. \cite{giorgi_relationship_2019} use LDA on annual reports and conduct a semantic network analysis to illustrate the connections and develop grounded theories of how firms interact with the legal environment. \cite{kaplan_double-edged_2015} use patent text to study the recombination in breakthrough innovation. \cite{sine_entrepreneurship_2022} study abstracts within a citation network. \cite{choudhury_machine_2021} use topic models to analyze CEO text and synthesize analysis of topic models using machine learning analysis of CEO facial expressions. 

\subsubsection{Limitation: Meaningless Output}

Latent themes that emerge inductively are meaningless and un-interpretable absent additional data (e.g., facial expression analysis to match themes (ibid)), or expert or crowd-sourced judgement. LDA assumes that \textit{all} of the text in \textit{every} document inside a text corpus is assigned to a theme. This is unwarranted if an expert has only a niche interest in a particular phenomenon. Papers we reviewed address this concern by limiting the scope of their corpus, or by extensively pre-processing to remove potential noise, but this also removes potentially relevant context and data. A second assumption of topic modeling approaches is that one must know the number of topics that exist within a collection of documents. This is non-trivial, particularly with massive corpora. 

Finally, topic modeling imposes an assumption that a researcher be interested in an \say{unbiased} representation. Experts often disagree about the interpretation of text, and a quest for unbiased (as opposed to transparent and replicable) analysis may be misguided \cite{nelson_measure_2019}. Ultimately, unsupervised computational methods face difficulty producing meaningful results in the absence of human involvement and expert interpretation \cite{nelson_computational_2020}. 

\subsection{The Modern ML Paradigm}
\label{sec:barriers}

Supervised machine learning (ML) models are widely used in many commercial applications: to evaluate candidates for employment \cite{cohen_midst_2022}, to support managers in giving employee feedback \cite{tong_janus_2021}, to moderate text content on the internet, judge the sentiment of customer-written reviews \cite{liu_assessing_2021}, and to build proprietary datasets for financial and academic use. Widely available ML methods can train a model to learn academic constructs and scale such models over vast volumes of unstructured text. However, computational models of text often struggle to provide insights that are interpretable, and may be perceived as too difficult for low-resource users to access.  Among 10 papers in the management corpus discuss ML, and four more that discuss AI, we could not locate any that describe the development of a new academic text classification model that is used to construct new data from unstructured text. 

While general AI and ML tools rapidly advance, there is a great need for applications in specific domains, such as detecting banking fraud \cite{abbasi_metafraud_2012} or fake websites \cite{abbasi_detecting_2010}. \say{Human-in-the-Loop} or hybrid systems are increasingly prominent.\footnote{Hybrid systems are not novel, with discussions already spanning decades. For example, placing a human in the loop was proposed in the design of \say{socially intelligent agents} \cite{dautenhahn1998art}. More recently, human involvement in security settings \cite{cranor2008framework} and cyber-physical systems \cite{schirner2013future} have appeared in the computer science literature.} The idea of human-in-the-loop for AI applications is promoted by \cite{zanzotto2019human} as a way to introduce responsibility into intelligent agents, as well as to increase interpretability. Likewise, human-in-the-loop for ML and NLP is particularly promising, both for data preprocessing and model learning \cite{wu2022survey}. Efforts to use AI and exclude domain experts are likely to fail, as shown in a study of how AI use in hiring decisions evolved into hybrid practices \cite{broek_when_2021}. Many AI systems cannot (and arguably should not) be trained without human oversight, and thus such intervention is vital to incorporating the necessary human knowledge into these systems.

Even with a human-in-the-loop approach, challenges have been noted, for example by \cite{xin2018accelerating}, which among others includes the ability to analyze the impact changes with iterative human intervention.  For academics in management and information systems, incorporating the knowledge of domain experts into the process of building structured data is essential for both deductive and inductive study. Such domain knowledge is crucial to forming the essence of training data, which necessarily represents some aspect of the worldview of a human who chose to train a model from it.

For supervised ML, challenges lie not so much with the ML methods or models themselves, but with the paradigm for ML. The process by which ML models are built (i.e. how information is learned) is opaque at the outset. Supervised methods in the Modern ML paradigm, and existing tools for the ML pipeline, do not give experts the ability to easily create training data and thus train a simple model that can scale over vast quantities of text. As discussed by \cite{Adadi2018}, much of the attention around Explainable AI (XAI) has been placed on \say{the way in which models behave,} seeking meaning and interpretability in the outputs of models that can often be highly opaque. If, instead, experts could easily build and curate training datasets, there would be greater opportunity to build interpretable models accepted by experts and members of the relevant audience. 

Labeled training data is the critical input to supervised ML models to code future, unseen text instances. What we call \say{the Modern ML Paradigm} for supervised learning of text data largely sees training data only as input, and holds the interpretability and explainability of ML model outputs as a post hoc concern. If building custom training datasets for text were easier, then resulting models can be interpreted -- and modified -- at the input stage. Expert-curated ML models could be interpretable and put into action in a domain, and escape some of the limitations of the modern ML paradigm: black boxes, proprietary training data, and un-reproducible, meaningless output that exclude participation from voices without supercomputers or advanced computer science knowledge.

\subsubsection{Limitation: The Black Box of Supervised Learning}
The \say{Black Box} refers to the mystery by which textual inputs are categorized by machines leading to certain outputs and are rooted in increasingly large model architectures and consequently, how computational models \textit{memorize}, rather than truly \textit{learn}. In turn, one is often hard pressed to decipher the behavior of a classifier, even with state of the art models -- they simply reflect patterns observed during training. 

Trust requires understanding \say{what the machines are doing} \cite{Castelvecchi2016}.  In the modern ML paradigm, attempts to \say{explain the black box} often come after the fact, which \cite{xai} calls \say{Post-hoc Explainability}. Here, when faced with models that are not interpretable \textit{by design}, one may try to boost explainability after the fact. As \cite{xai} writes, post-hoc approaches have become the \emph{de facto} way of achieving some level of explainability with complex DL models. 

Without understanding the data that goes into training the model, the explainability of the modern ML paradigm is limited. In a recent survey of explainability in supervised ML methods, \cite{Burkart2021} call for rethinking the problem from first principles, and encourage researchers to ask \say{What are we actually looking for? Do we really need a black box model?} For many experts, the immediate answer is no. Even a ML model with a high degree of predictive power is likely to be seen as undesirable if it does not produce interpretable results \cite{nelson_measure_2019,nelson_computational_2020}.

\textit{Underspecification} can result from training models on massive volumes of data and selecting the model(s) that achieve a desired predictive power \cite{damour_underspecification_2020}.  Underspecification often leads to failure when those models are used outside of the context in which they were trained. As \cite{heaven_2020} writes: \say{the process used to build most machine-learning models today cannot tell which models will work in the real world and which ones won't.} \cite{doi:10.1126/scirobotics.aay7120} calls for greater interaction between humans and machines and for explainability when models are faced with new situations -- a \say{sign of mastery.}

\subsubsection{Limitation: Proprietary Training Data and Methods}

Seen from a different angle, black boxes may be valuable in order to protect trade secrets hidden behind such trained models \cite{https://doi.org/10.48550/arxiv.1811.10154}. Secrecy presents risks in the cases where \say{high-stakes} decisions are made by the models in question, having potentially significant societal or economic implications. Academic users often rely on commercial data providers that give little to no insight into their proprietary process for data creation. In our analysis of the management literature, 13 of the 100 text analysis papers, and 117 papers in total indicate the use of licensed, proprietary data or models. The problem identified by \cite{lazer_computational_2009} is that such computational methods could become the \say{exclusive domain} of private companies and government agencies that operate contrary to the academic commitment to openness.  The inadequacy of data-sharing paradigms for big data cast doubt regarding the veracity of ML models \cite{lazer_computational_2020}. 

Private firms and academics typically advance the state of the art by breaking new records for the complexity of data, the size of models, and the number of parameters in a model. Many potential contributions from researchers  who lack access to super-computing resources, the funds to buy large proprietary datasets, or access to confidential government data are excluded \cite{card_expanding_2010}. Our search for management papers that indicate proprietary data or models illustrate stable trends over time, but when big or secret data outputs become the criteria through which the quality of research is judged, only those with access can compete, and there may be less attention to \say{cumulative progress toward answering important questions} \cite{davis_editorial_2015}.

By reducing the need for humans to think, act, or belong (contribute, be included) in the public sphere, \cite{kane_avoiding_2021} write that ML systems can constrain and oppress, rather than achieve emancipatory goals. To deliver on the potential of ML, improvements needed include greater sharing of data, protections for privacy to enable more open data, intellectual property standards to reduce transaction costs, and approaches that enable genuine replication \cite{king_gary_ensuring_2011}. Pre-processing of datasets for supervised learning could be a core strength and contribution of social scientists in analyzing text data \cite{dimaggio_adapting_2015}.\footnote{One promising approach is computational grounded theory: a novel framework to use unsupervised machine learning to identify patterns, engage a human in interpretive refinement of patterns, and use computational methods to assess and confirm identified patterns \cite{nelson_computational_2020}. This \say{human-centered computational exploratory analysis} is proposed to help social scientists benefit from unsupervised computational methods and enhance the transparency and reproducability of content analysis \cite{nelson_computational_2020}.} Despite  advances, DiMaggio's (2015) call for greater systematic research to develop solutions for curation challenges and guidelines for pre-analysis and developing training data has largely been unmet.

%In particular, this essential human knowledge is incorporated in CRAML into the phase of rule creation, which becomes the foundation upon which the resulting datasets are crafted. 
\subsubsection{Limitation: Meaningful Contributions to Knowledge}

In a statement that rings true almost a decade after it was written, there is little evidence of computational social science in leading social science field journals \cite{watts_computational_2013}, or in the management corpus reviewed here. Computational social science methods have advanced significantly, but it has yet to be demonstrated how computational scientific infrastructure can be applied to important social problems \cite{lazer_computational_2020}. Disciplinary silos, privacy concerns, and unreliable, non-replicable, and proprietary data hamper the potential for computer science methods to provide meaningful insight into society's challenges \cite{lazer_computational_2020}. 

The emergence of a computational social science apart from traditional academic disciplines speaks to the hope that ML and AI can advance knowledge. However, the possibilities of these technologies are lessened when such tools are perceived as beyond the reach of ordinary users \cite{faik_how_2020}. Less than a decade old, some such initiatives, such as \say{data science for good}, have been criticized for not giving the social sector \say{the opportunity to design for what we want} \cite{porway_funding_2022}. Without tools that enable domain experts to develop their own ML models, there is a risk that new interdisciplinary or trans-disciplinary centers might lead to yet another siloed academic discipline as ill equipped to solve problems as others \cite{abbott_chaos_2001}. Future opportunities for AI identified by \cite{benbya_artificial_2020} include democratization, reducing requirements for data, and enhancing AI explainability and transparency.\footnote{Emerging paradigms for low-resource Natural Language Processing (NLP) seek to eliminate barriers to accessing ML technology \cite{hedderich_survey_2021}.}

\cite{https://doi.org/10.48550/arxiv.1206.4656} attributes a \say{Hyper-Focus} in current ML research to three causes that have led it to lose \say{its connection to problems of import to the larger world of science and society.} First, the (over)use of benchmark datasets, many of which are \say{synthetic}, has \say{glaring limitations} and still make reproducability difficult. Only 1\% of ML papers are applied to a specific domain and the rest use benchmark datasets.\footnote{Classic examples of these datasets include the Movie Review Dataset \cite{maas-EtAl:2011:ACL-HLT2011} or the Yelp Review Dataset. While use of standardized datasets may reduce the reproducibility crisis in ML \cite{kapoor_narayanan}, there is a tradeoff. As one might imagine, benchmark and standardized datasets have limited application to the real world; furthermore, many benchmark datasets are tailored for specific tasks, e.g. movie reviews are often used for sentiment analysis.} By way of contrast, \cite{larsen_tool_2016} report that 90\% of research constructs are uncommon, suggesting that using pre-labeled standard benchmark data for training will miss entire vast areas of domain-specific knowledge. Secondly, the focus on \say{abstract metrics} like F-scores to determine model quality across domains is a \say{mirage} that masks the need to focus on impact and usefulness of a model. A final issue is the \say{lack of follow-through,} suggesting that current computational research leaves little incentive for connection to sustained research agendas.

A related concern is the rapidly depreciating relevance of data that many ML models are trained on \cite{lazer_computational_2020}. In the ML training process, models are trained \say{in the lab} on a training dataset, yet the environment in which models are deployed is dynamic. \say{Big data} research may apply only to specific users of a specific portal at a specific time in which a study is conducted  \cite{hargittai_is_2015, pfeffer_this_2022}. For that reason, online studies and experiments with large response rates may fail to produce reliable or general knowledge \cite{bradley_unrepresentative_2021}. A static training process may simply not reflect the real world environment or its rapid evolution \cite{babic_cohen_evgeniou_gerke_2020}. The velocity of internet-based data also creates a limitation for computational social scientists' conclusions \cite{munger_limited_2019}. Solutions to these obstacles include forming new multi-disciplinary, international journals that permit rapid publication of rigorous quantitative descriptions of rapidly evolving phenomena \cite{munger_quantitative_2021}.

\subsection{Seeking Explainable Tools for Social Science}

To sum up the issues described in this section, each method reviewed has strengths, limitations, and tradeoffs for experts who seek to analyze unstructured text. For domain experts with an interest in specific phenomena, the methods reviewed so far do not allow for expert-led, targeted extraction of niche constructs hiding within corpora and the construction of standardized data. For many qualitative researchers, the systematic analysis and discovery of new patterns within domain-specific digitized records is a core contribution -- the ability to find the diamond (a single topic) in the rough (lots of text). If qualitative hand-coded contributions could be easily scaled, it can address some issues found in the supervised ML domain.

%% file: motivation.tex
\section{Context Rule Assisted Machine Learning (CRAML)}
\label{ref:motivate}
Enabling experts to create training data for niche classifiers to detect constructs that are relevant within a specific domain would address many of the problems identified with the modern ML paradigm, and the limitations of unsupervised computations methods. The CRAML method we present emphasizes explainability and has a contribution on each of the six dimensions that explain the performance of AI systems according to \cite{asatiani_challenges_2020}: the models are specific to a single construct, the goals are clear, by limiting the chunk size and extracting text from documents in the corpus, the training and input data is intended to be context-specific, the output data is interpretable (a 0/1 binary classification), and the environment is also intended to be domain specific and expert-guided. By emphasizing the ``human in the loop,'' CRAML gives qualitative and quantitative researchers control over the training data that is the key input to ML models. This offers new capabilities to control the analysis and measurement of phenomena found in massive volumes of unstructured text.  This section briefly introduces the solution developed and explains how the solution addresses the problems described above. 

%To confront a specific challenge arising from a needle in the haystack problem with a very large text corpus, we developed a human-centric, hybrid system to bridge the gap between unstructured text data and the modern ML paradigm. We aim to harness the explainability and trust ultimately gained from manual and transparent codifying of text by domain experts, ease the process of creating novel training data, and enable experts to steer learning with ML models that can scale expert-created schema over vast amounts of unstructured text in a probabilistic manner.

\subsection{A Framework for Unstructured Text Analysis}

CRAML software gives users the tools to manipulate keywords to extract relevant data from massive corpora, create \say{tags} that seek to capture topics or themes, and subsequently, develop \say{context rule} sets that codify knowledge about the specific pieces of text that correspond to specific tags. The CRAML process yields tabular output that matches document-level metadata with a 0/1 indicator that symbolizes the occurrence of tags in a document \cite{lazer_data_2017}.

\begin{figure}[ht]
\centering
\includegraphics[scale=0.6]{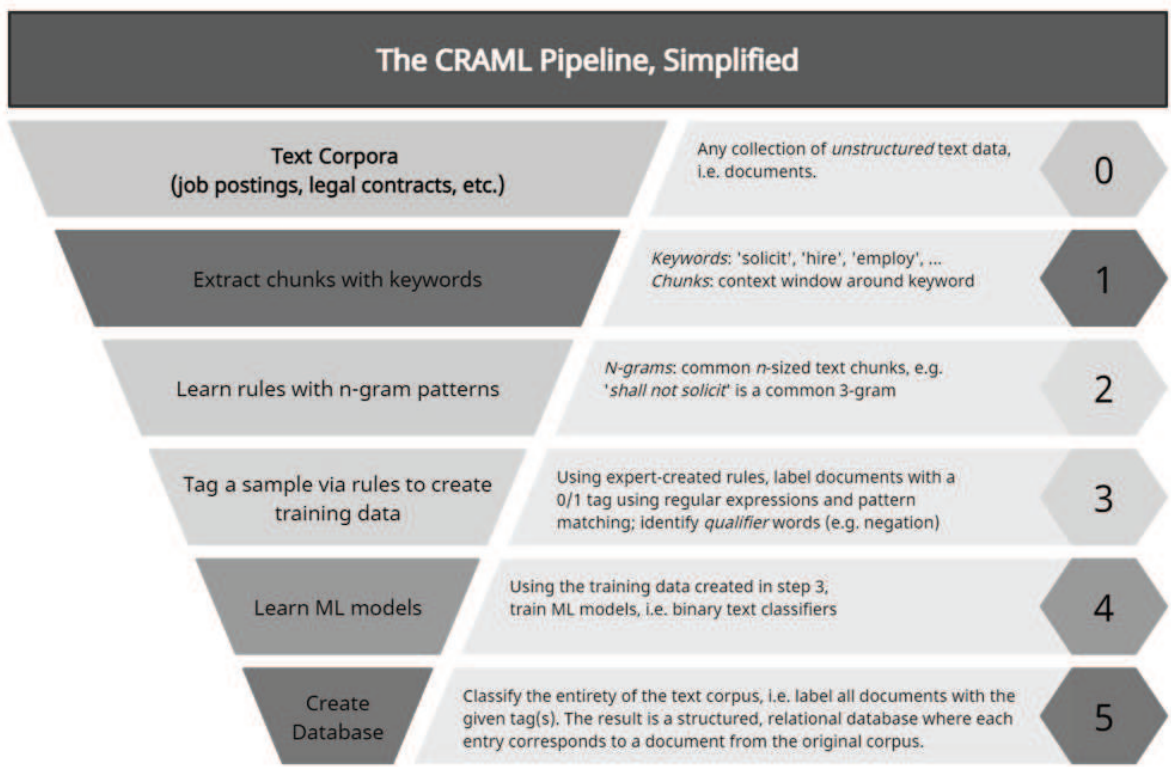}
\caption{A Simple View of the CRAML Process}  
\label{fig:simple-craml-process}
\end{figure}

Figure \ref{fig:simple-craml-process} provides a high-level overview of the CRAML framework. The CRAML process begins with the full text of a corpus of documents (Step 0). While metadata is not required, CRAML software is designed to combine characteristics extracted from the text with metadata at the document or record ID level.

In step 1, the researcher chooses keywords to extract from the corpus, and retrieves a \say{\emph{chunk}}: a block of text containing a user-selected number of words surrounding the keyword. The chunk length should contain the full \say{\emph{context window}}: all of the relevant context for a human to determine if the keyword-containing chunk is relevant or not for some binary classification schema. The user can extract all chunks from all of the documents in the corpus, or in low-resource environments, can randomly sample from the documents in order to perform Steps 2-4 on a smaller subset.

In step 2, the researcher analyzes the patterns within the chunks, studying n-grams that reveal the most common phrasings surrounding the keyword. 

In step 3, the researcher writes \say{\emph{context rules}} that elaborate an ever-more detailed scheme of binary classifications of the chunks. We refer to these binary classification schemata as \say{\emph{tags}}: each tag represents a topic or theme that the researcher defines by recognizing patterns in the extracted text and determining if each chunk is (tag=1), or is not (tag=0), illustrative of the tag. There is no limit on the number of tags a researcher can define. For each tag, there is no limit to the number of rules that can be defined using either exact text matching or regular expressions (RegEx).

In step 4, the researcher written context rules are applied or \say{\emph{extrapolated}} over either the full extracted chunks or a sample of the extracted chunks to create a training dataset for ML modeling. Extrapolation labels each chunk with a 0 or 1 for each “tag,” or binary classification scheme, and creates a dataset that can be used for training ML models. In step 5, common ML algorithms use the training dataset to tag the full extracted chunks, and the researcher identifies the best performing model. In the final step, the results are aggregated into a document-level database of structured information: for any document within the text corpus, the researcher knows if the document contains a tag. Additional hand-coding by independent third-parties can validate that the training dataset extrapolation, and the ML results, are highly accurate.

\paragraph{Versatile tools for mixed approaches}
CRAML's pipeline incorporates a novel set of tools that provide an end-to-end solution to the challenge of structuring unstructured text data. However, researchers may appreciate its flexibility and multiple uses: a researcher may wish to pursue analysis using only pieces of the CRAML process. The proposed system is not a rigid algorithm, but rather a flexible pipeline of tools created to implement a research framework that was developed to address a specific research task. As \cite{pandey_applying_2019} note, choosing between a dictionary, rules, or supervised machine learning involves tradeoffs, and no clear guidelines. In CRAML, particular steps and technologies can be interchanged according to the user's preference, skill set, and research challenge. A qualitative researcher can quickly sample and sift through  text data and experiment with alternative classification schemes and definitions.\footnote{Emerging paradigms for low-resource Natural Language Processing (NLP) seek to eliminate barriers to accessing ML technology \cite{hedderich_survey_2021}. Particularly with low-resource NLP, the availability of labeled data, i.e. text, become crucial.} For another researcher, extraction could be a pre-processing step before engaging in a computational grounded theory project that uses unstructured NLP methods once the plausibly relevant themes are extracted based upon keywords. A third researcher may wish to extrapolate a full dataset without ML. A fourth researcher may wish to only sample a massive corpus, extract context windows around keywords, and inductively generate topics or themes with extracted text and have no desire to perform further computational avenues of analysis.

%\subsection{An Expert-Centered Approach to Steer Machine Learning}

%The modern ML paradigm assumes \emph{accurate}, structured, labeled training data as an input, but excludes the question of how to build the training data from unstructured text.

\subsection{A Bridge to Solving Real-World Problems}

By positioning context rules as the crucial bridge between domain expert knowledge and structured datasets that represent embodied knowledge, CRAML provides a \say{traceability} from datasets classified by ML models back to the context rules that lead to them -- and the humans who wrote those.  By placing a focus on explainability at the first bridge from unstructured data to structured data(sets), the second bridge from training data to trained models becomes easier to cross. ML modeling ``presumes a world of already existing (divine or rationalistic) rules, which only need to be formalised, in order to make sense to a machine (or an analytical philosopher for this reason)'' \cite{Apprich2018}. Context rules in CRAML codify which text belongs to which topic or theme according to the user. To the extent such rules can be articulated consistently, then they can be extrapolated over unstructured text to create fully structured datasets to train ML models. 

Rather than presuming unbiased training data, CRAML extends the user's worldview to the realm of ML models via expert-created context rules. Interpretive frameworks lend meaning to text and are a cornerstone of academic research. Scaling methods such as coding and grounded theory  over large volumes of text contributes to theory development \cite{tchalian_microfoundations_2019}. Rigorous manual content analysis by qualitatitve social scientists is often geared toward the induction of topics and themes, yielding new constructs, analytical frameworks, or mid-level theory of some phenomenon \cite{glaser_discovery_1967}.\footnote{For scientific purposes, inductively derived analytical frameworks from text should be a) intra-subjectively reliable, b) inter-subjectively reliable, and c) fully reproducible \cite{nelson_computational_2020}. This means that a single person should reach the same coding conclusion again and again, a second person should reach the same conclusion, and the methods, data, and analysis should be transparent enough to permit replication.} Alternatively, a quantitative or hypothesis-driven researcher might pursue evidence of a phenomena within a text corpus, quantify its frequency, and then analyze it. 

In CRAML, the use of context rule sets serves as the foundation for representing the knowledge of a domain expert, without the need for explicitly annotating all unseen instances (i.e. documents). Through the extrapolation process, the hybrid CRAML framework requires manual work only in the crucial first stages, and leaves the remainder to automation. CRAML's initial step is similar to processes used in IS research: for example, to develop a specialized ontology of hypotheses,  \cite{li_theoryon_2020} couple rule-based approaches for extraction and word embeddings. The ``human in the loop'' contributes transparent manual codifications of text, which allow for the replicable and configurable construction of tabular datasets used either immediately for quantitative social science analysis, or to train ML models.\footnote{Explainable and interpretable models can be built when users ``steer learning'' through feedback and interaction \cite{doi:10.1126/scirobotics.aay7120}. \cite{Adadi2018} writes that \say{explainability can only happen through interaction between human and machine.} Of course, the creation of such hybrid system requires expert human labor.  \cite{https://doi.org/10.48550/arxiv.1811.10154} writes that \say{[i]nterpretable models can entail significant effort to construct, in terms of both computation and domain expertise.}}  Because CRAML involves human expertise and gives control over training data to the user, the ML model that results from CRAML is interpretable as a product of a human-built and curated rule set that can be published, and then contested, examined, and modified by others. This mitigates  concerns over black box ML models. 

CRAML built models can have real-world applicability, because training data and ML-built datasets can be formed by domain experts who largely address Wagstaff's (\citeyear{https://doi.org/10.48550/arxiv.1206.4656}) concerns regarding abstract metrics and lack of follow through. Empowering users to create novel, meaningful datasets and ML models in their worldview also enables rapid changes that can be put into action locally. This comes at a time where novel ML or DL approaches can be severely limited by available datasets, with no widely accepted approach to create novel ones as needed and in an efficient manner.

As a final response to the challenges introduced in the previous section, CRAML's methodological transparency is intended to address several of the problems with proprietary data and otherwise unreachable methods. With a standard desktop computer and open source software, CRAML enables interactive engagement with massive text corpora without prior programming knowledge (although users will find some programming knowledge helpful and advanced users may seek to customize the underlying code for niche purposes).  At each crucial step of the framework, CRAML saves and shows the user the underlying data in CSV format in order to enable the interactivity essential to truly achieving a \say{human in the loop} and records each step of the process for replication and transparency purposes. 

By making the framework as open and transparent as possible, and publishing the code base and the underlying data used here, we provide opportunities for users to test, harness, or modify the framework. CRAML software is available under a  under a Creative Commons Attribution-NonCommercial-ShareAlike 4.0 International License. The artifacts created and used by CRAML in the intermediate stages are extracts and samples of the plausibly relevant text from a document and document corpus using keywords and context windows. A user can control what information is extracted from the corpus, and thus remove sensitive data. Thus, CRAML can mitigate some privacy protection and sensitive information concerns that limit access to sensitive data. Other researchers working with big data have taken a similar approach by extracting only relevant pieces of larger documents \cite{suarez_entrepreneurship_2021}. By 

\subsection{An Application of CRAML: Job Advertisement Text}

Job seekers face difficulties finding the right jobs, and employers face frictions in job search that represent additional recruitment costs. Search frictions harm both workers and employers by increasing the barriers to good matches in the labor market, and reducing competition for labor \cite{burdett_wage_1998,manning_monopsony_2003}. A suite of ML classifiers that added niche filters to job search engines could benefit jobseekers, employers, and workforce agencies. With support from the National Science Foundation, the National Labor Exchange created the NLx Research Hub in 2021 to ``increase the amount of actionable labor market information in the U.S. to facilitate the recruitment, hiring, and training opportunities of American workers'' and ``deepen partnerships between industry, government, and academia by enhancing the infrastructure to support the convergence of research, education, and talent pipelines.''

Workers who desire more flexible, remote jobs after COVID-19, disabled workers seeking specific working conditions, union workers, workers with a professional license, and veterans as well as veteran’s spouses may all be underserved even while being preferred by specific employers. An employer can gain a competitive advantage by hiring workers who are otherwise discriminated against, but only in an efficient and competitive labor market \cite{becker_economics_1957}. Job seekers may easily become discouraged without information on whether, for example: a job is located near public transit, or whether an employer will interview an ex-felon, a teen, or a non-resident on a visa. If some employers seek to target and hire underserved audiences, but job ads never reach the workers, then research should aim to reliably identify which jobs might be especially relevant for a particular worker. 

\paragraph{Research application}
CRAML was first developed using job advertisement text data to support research on the labor market. While the underlying text is confidential and licensed, we describe and release rules files and nine ML classifiers that achieved a high accuracy in  detecting job characteristics and barriers to employment under a  Creative Commons Attribution-NonCommercial-ShareAlike 4.0 International License. The research implications of this work are not only immediate research papers, but a potential to release custom-built ML classifiers and create open tools for real-time information systems that track changes in the labor market. Such application could improve both research and contribute to reducing barriers to employment in job advertisements. Ultimately, better data on employment practices obtained from job advertisement text and other sources of text information could have a profound impact on research capabilities in this field.\footnote{In ongoing work supported by the Russell Sage Foundation Future of Work Program and NLX, a job-advertisement level SQL database built from CRAML's classifiers using NLX data is now operational and  new classifiers are under construction.}

In Appendix \ref{sec:appendixb}, we provide a list of tags and the keywords extracted as part of an effort to identify specific barriers to employment, the accuracy of the models built, and discuss how the broader labor market research community can contribute to this effort.

%% file: data_analysis.tex
\section{A Step-by-Step, Replicable No Poach Classifier}
\label{sec:analysis}
We demonstrate the start-to-finish CRAML approach by applying it to an empirical context where the resulting analysis might be meaningful, transparent, and replicable. Toward this end, we contribute a new text corpus of 151,708 franchise documents, and the cleaned and pre-processed text we used in this analysis. We publish metadata that tracks which of 12,992 records a particular document corresponds to, and includes the effective date, company name, and unique ID. We also publish the rule sets that are the manual hand-coded input, enabling replication and scrutiny of the results. We also publish the training dataset and the ML model.\footnote{The PDFs are hosted in partnership with \protect\href{https://documentcloud.org/app}{DocumentCloud}, and will be released upon publication. On DocumentCloud, where they can now be searched, discussed, and annotated. Additional items will be uploaded to a scholarly repository upon publication.}

\subsection{Empirical and Data Context}

``No poach'' clauses in contracts are also referred to as collusive pacts or anti-competitive restraints on employee mobility, and have drawn interest from academics, regulators, and policy-makers. While anti-competitive language is contained in public documents, sometimes contrary to public policy, these documents are inaccessible for many: available only in massive, unsearchable collections of PDF documents on state agency websites. With subtle variations in language and no guide about where to look for these clauses, no poach clauses are diamonds in the rough: a few sentences in hundreds of thousands of documents with millions of pages. Indeed, such clauses were largely not known to exist by the relevant academic and policy community of interest  until the release of a 2017 working paper published as \cite{krueger_theory_2022}. The paper contained a limited sample and relied on a third party data provider to identify no poach clauses.  Following the release of the working paper, eleven state attorneys general issued a letter in 2018 demanding the practice end,  many franchises voluntarily ended the practice, and the State of Washington in 2018 first negotiated settlements with franchise companies in which many removed their no poach clauses.\footnote{See \protect\href{https://agportal-s3bucket.s3.amazonaws.com/uploadedfiles/Another/News/Press_Releases/NoPoachReport_June2020.pdf}{Washington State Attorney General Report}, \protect\href{https://attorneysgeneral.org/wp-content/uploads/2019/02/2018.07.09-No-Poach-Agreements-Letter-Redacted.pdf}{Letter from State Attorneys General}. As Ashenfelter wrote, ``it is instructive that the mere revelation of collusive agreements, whether legal or not, has so quickly provoked a strong response from both the antitrust authorities and the franchisors whose agreements contained these no-poach clauses'' (qtd. in \protect\cite{krueger_theory_2022})}.

Both no poach and non-compete clauses are known to place downward pressure on wages \cite{callaci2022effect,balasubramanian_locked_2022}. Given their importance in reducing opportunities for job mobility, there is great interest in descriptive statistics regarding the prevalence of these clauses. Overcoming the inaccessibility of the text within public documents was a significant challenge. We assembled a large collection of Franchise Disclosure Documents (FDDs) from the state of California by scraping PDF documents from the public web. These records exclude companies that may be exempt from filing, and may be missing observations due to limitations when accessing data from the web.\footnote{While we were able to obtain and match the 151,708 documents to the 12,992 records via web scraping, we identified 16,216 franchise disclosure records from January 2013-July 2022, and are investigating additional methods to acquire any missing data, which we believe to be randomly distributed.}

%We assembled a large collection of Franchise Disclosure Documents (FDDs) from the states of California and Minnesota by scraping PDF documents from the public web.

In addition, we make available the machine-readable text corpus we build from the downloaded PDFs and the metadata that ties documents to specific company filings. For pre-processing, we used Python's Tika, and for files that could not be initially read with this, we used ABBYY Fine Reader. For the California corpus, we assembled 151,708 documents in 6.99 GB of cleaned machine-readable text that can be traced back to a total of 12,992 franchise records.  We consider the record to be the unique ID that the state assigns to multiple documents from a franchise on a particular date. We trace each document back to a record with metadata that includes the name of the franchise, and the date of the filing.

%For the Minnesota corpus, we assembled 5,493 documents in 1.64 GB of cleaned text that can be traced back to 4,201 franchise records. 

%The California corpus was larger, so we focus on it for primary analysis. 

\subsection{The Construction of Training Data}

\subsubsection{Step 1: Extract Relevant Text to Reduce Information Overload}

\begin{figure}[htbp]
    \centering
    \includegraphics[scale=0.55]{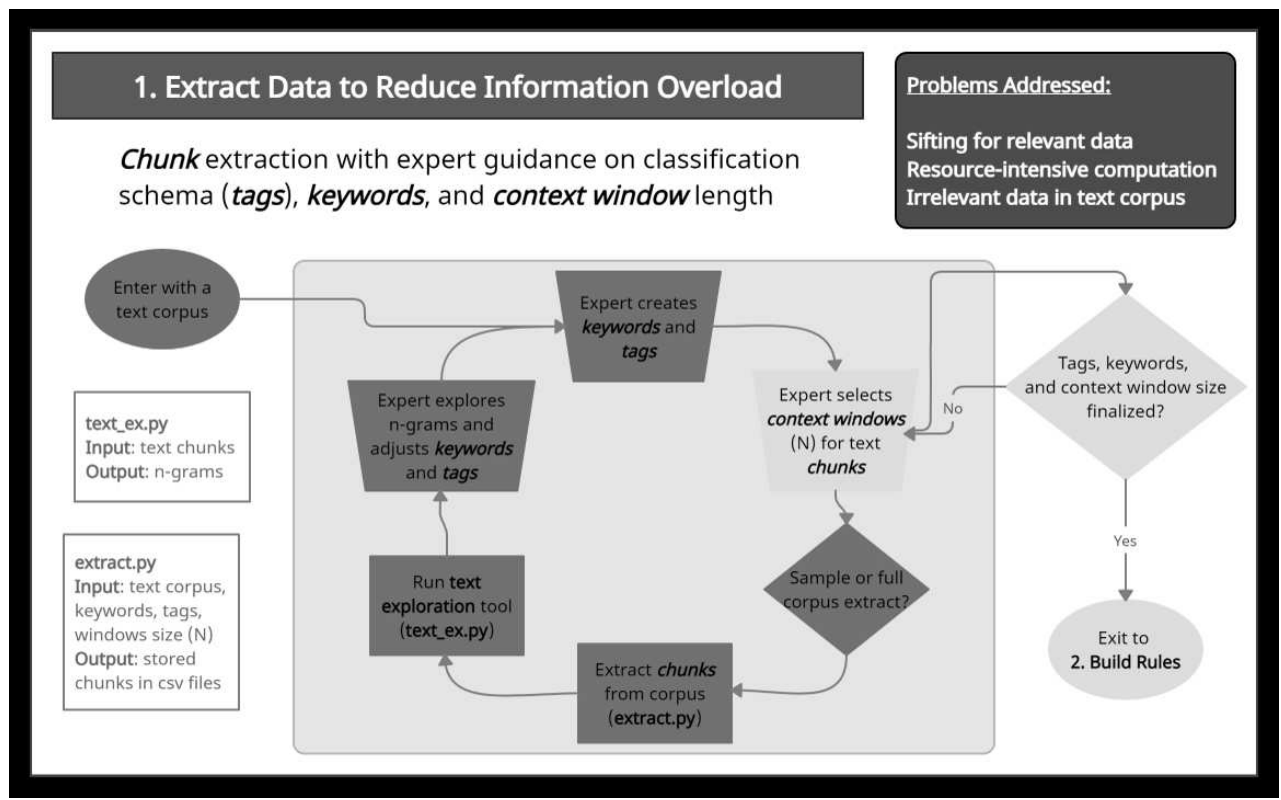}
    \caption{First process in the CRAML Framework -- Extract Data}
    \label{fig:extract}
\end{figure}

The extraction process encompasses taking the unstructured text files and preparing a cleaned dataset of only the text that is plausibly relevant that is ready for hand-coding. By reducing the size of the text files being worked with, and delimiting the extracts to the relevant text in the corpus, this step addresses challenges of computational resource intensity, and assists in sifting for relevant information in a large corpus. This is a manual, iterative process -- a loop that is exited when the user is satisfied that all of the relevant text is extracted from the corpus. Figure \ref{fig:extract} illustrates the extraction process, which corresponds to Step 1 outlined in Figure \ref{fig:simple-craml-process}. Two main software tools (extract.py and text\_ex.py) aid the user in finding and exploring the relevant text.

We chose initial keywords \say{hire}, \say{recruit}, \say{employ}, and \say{solicit} to explore if franchise documents contain a no poach clause. The initial keywords chosen suggested additional ideas for tags, which in turn, suggested new ideas for keywords. To aid in this process, a repeated N-gram exploration helped discover the contexts in which given keywords appear the most. This exploration prioritizes the extract, suggest new ideas for keywords, tags, rules, context windows. Because the full context was important to explore initially, the earliest exploratory extracts focused on the 20 words surrounding each keyword. Following all of the processes described below, a final context window of 13 words was selected (6 words to the left and right of the keyword). The goal when removing irrelevant or non-essential information is to reduce noise and build a highly focused ML classification model. In an iterative and flexible process, more keywords were subsequently added: for example, ``poach'', ``non compet'', ``noncompet'', ``covenant'' and ``not to compete'' simply restarted the process. 

\subsubsection{Step 2: From Rule Sets to (Training) Data via Extrapolation, Testing, and Validation}

\begin{figure}[htbp]
    \centering
    \includegraphics[scale=0.55]{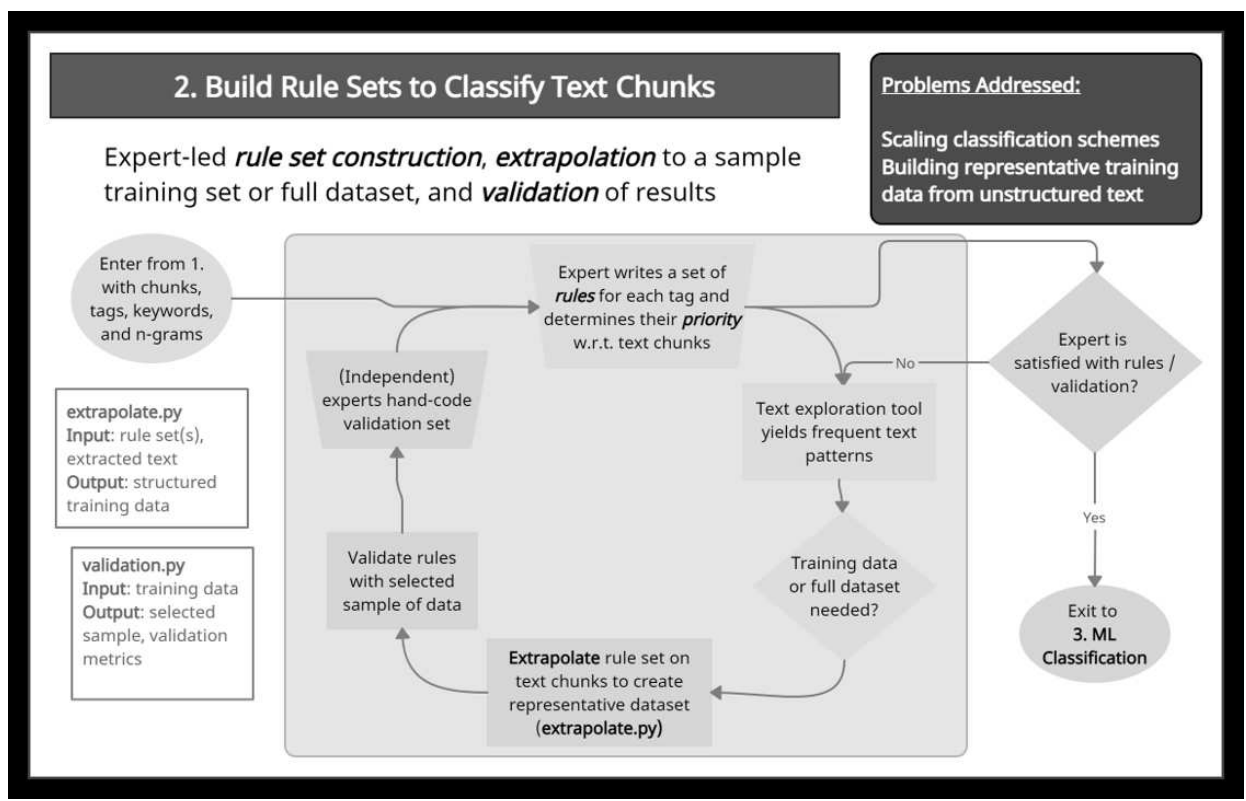}
    \caption{Second process in the CRAML Framework -- Build Rules}
    \label{fig:rules}
\end{figure}

After the extraction process, an expert conceptualizes and defines the desired classifications, or \textit{tags}, and builds a  rule set for each tag in a manual and iterative fashion. Figure \ref{fig:rules} details the iterative process of Step 2 in Figure \ref{fig:simple-craml-process} that involves validation, extrapolation, and the refinement of rule sets.  This process concludes when the expert is satisfied that the extrapolated rules are valid on a sample of the text. The rule set is then ``extrapolated'' to the full extract from the corpus, yielding a structured dataset of accurately classified chunks -- a training dataset, in other words. As described in the following section, the results of Step 2 can be used immediately for descriptive research regarding the contents of documents, or as a basis for training ML models.  

The cycle begins with the initial creation of rule sets for each tag, which can be done manually by an individual researcher, or team of researchers. Constructing and designing accurate indicators requires researcher familiarity with ontological subtleties and appropriate construct validation and scale development methodology \cite{weber_constructs_2021,mackenzie_construct_2011}.  Rules are written in a CSV file that contains a list of rules, the priority level of each rule, and the corresponding tag, assigned a 0 or 1. As earlier, the N-Gram tool may be helpful -- doing this prioritizes the creation of rules that will \say{cover} the largest parts of the entire data. Priority level determines the order in which a rule is run. A earlier rule can be over-written by a subsequent priority rule. 

Step 2 then involves the use of rule sets and the extrapolation algorithm (extrapolate.py) to validate the rules. In essence, the extrapolation process will convert the set of rules into an extended dataset, and will tag each chunk according to the user's encoding of each rule. The resulting dataset contains the unique document identifier, the extracted chunk matching a certain rule, and the defined encoding for this chunk. Note that this extrapolation is done individually for each rule set and tag, thus resulting in a training dataset for each rule set (file).  A second software tool (validate.py) relates to the validation of the data that is the output of extrapolation. Once extrapolated, the expert tests the performance of the rules against actual chunks that are coded by those rules. The expert can export a strategic sample of chunks meant to ensure each rule is valid -- choosing a random sample of N examples per rule. This enables independent hand-coders to score each chunk, which can then be re-imported to assess accuracy and performance of the rules.

For the franchise document data, allowing iterative exploration of keywords and adjustments to rules, we developed a set of over 500 rules and 6 tags to classify suspect anti-competitive clauses in the corpus. We first focused on finding all employee ``no poach'' clauses. As one rule for the no poach tag states: ``\textit{you may not seek to employ or retain any employee or independent contractor who is at any time employed by us.}'' When examining similar extracted chunks, we found that many documents also contained language that went further than prohibiting an employer from soliciting another party's employee, but also prohibited an employer from hiring or employing. We create a separate tag these ``no hire'' clauses. Such ``no hire'' clauses, as in the following example from the no hire rule set, state that: ``\textit{party shall not hire or solicit to hire any person employed then or within the preceding year by the other party.}'' 

%Because we knew that many franchises voluntarily ended their practice of enforcing no poach or non compete covenants, we searched for limits to anti-competitive practices found within the text and created a third tag: ``limitations.'' One limitations tag rule states that the franchiser must ``\textit{not restrict restrain or prohibit a franchisee from soliciting or hiring any employee}.'' We also noted that some of the language limiting restraints upon employee mobility also forbade ``covenant not to non compete'' clauses. For example, another rule of for the limitations tag states that such a restrictive ``covenant is void and unenforceable against an employee.''

Independent validation in which a third researcher hand-coded 706 chunks indicates an initial 91\% match between an independent third party and the extrapolation of rules, suggesting a high degree of inter-rater reliability in ability to detect characteristics of no poach clauses. While our emphasis here is to ensure that there is strong inter-rater reliability between the CRAML user and an independent observer in this test example, we do not claim perfect identification of all no poach clauses. For a regulator, additional scrutiny would be required, but the overarching goal is to identify plausibly relevant and/or problematic no poach language. The pipeline enables any observer to classify the documents according to any arbitrary schema. We emphasize that CRAML here enables content validation processes at the input stage to ML. Independent assessments of reliability are key to any conclusions. 

\paragraph{Step 2 Results: A Labeled, Structured Dataset for Analysis}

Step 2 yields a dataset for training a ML classification model, or, if the underlying corpus is small enough (as is the case here), the rules can be applied to the entire dataset and reveal which filings contain which clauses according to the specified rules. 

%Before returning to the CRAML process below and describing the creation and performance of the resulting classifier, we present several figures obtained from the extrapolation of rules to the full dataset of franchise documents. These represent an exact match between the rule set and the text within a document and a record.

%\begin{figure*}[ht]
%    \centering
%    \begin{subfigure}[b]{0.475\textwidth}
%            \centering
%            \caption{California Corpus}
%            \includegraphics[width=\textwidth, scale=0.2]{data/ca_rule_vs_ml.png}
%            \label{fig:combofiga}
%    \end{subfigure}
    %\hfill
    % \begin{subfigure}[b]{0.475\textwidth}
    %         \centering
    %         \includegraphics[width=\textwidth, scale=0.2]{data2.png}
    %         \caption[]
    %         {{\small Non Compete, Narrow, and Jurisdiction Clauses}}    
    %         \label{fig:combofigb}
    % \end{subfigure}
    % \hfill
    % \begin{subfigure}[b]{0.475\textwidth}
    %         \centering
    %         \includegraphics[width=\textwidth, scale=0.2]{data3.png}
    %         \caption[]
    %         {{\small Naked No Poach and Non-Compete Clauses}}    
    %         \label{fig:combofigc}
    % \end{subfigure}
    % \hfill
    % \begin{subfigure}[b]{0.475\textwidth}
    %         \centering
    %         \caption{Minnesota Corpus}
    %         \includegraphics[width=\textwidth, scale=0.2]{data/mn_rule_vs_ml.png}   
    %         \label{fig:combofigd}
    % \end{subfigure}
%    \caption{Analysis of Suspect No Poach Clauses in California Franchise Documents} 
%    \label{fig:combofig}
%\end{figure*}

\subsubsection{Step 3: Training ML Classifiers to Build Structured Datasets}

While the above analysis is based on an extrapolation of rules to all chunks in the dataset, CRAML yields data that a ML classifier can train. Although many algorithms exists for such a training process, the CRAML framework trains a binary classifier for each tag. In other words, each training dataset is created from a rules file, which contains one or more tags. Accordingly, each classifier can be traced back to a rule set to perform a 0/1 classification for the tags included therein. The training process is described in greater detail in  Appendix \ref{sec:appendixc}.

\begin{figure}[htbp]
    \centering
    \includegraphics[scale=0.55]{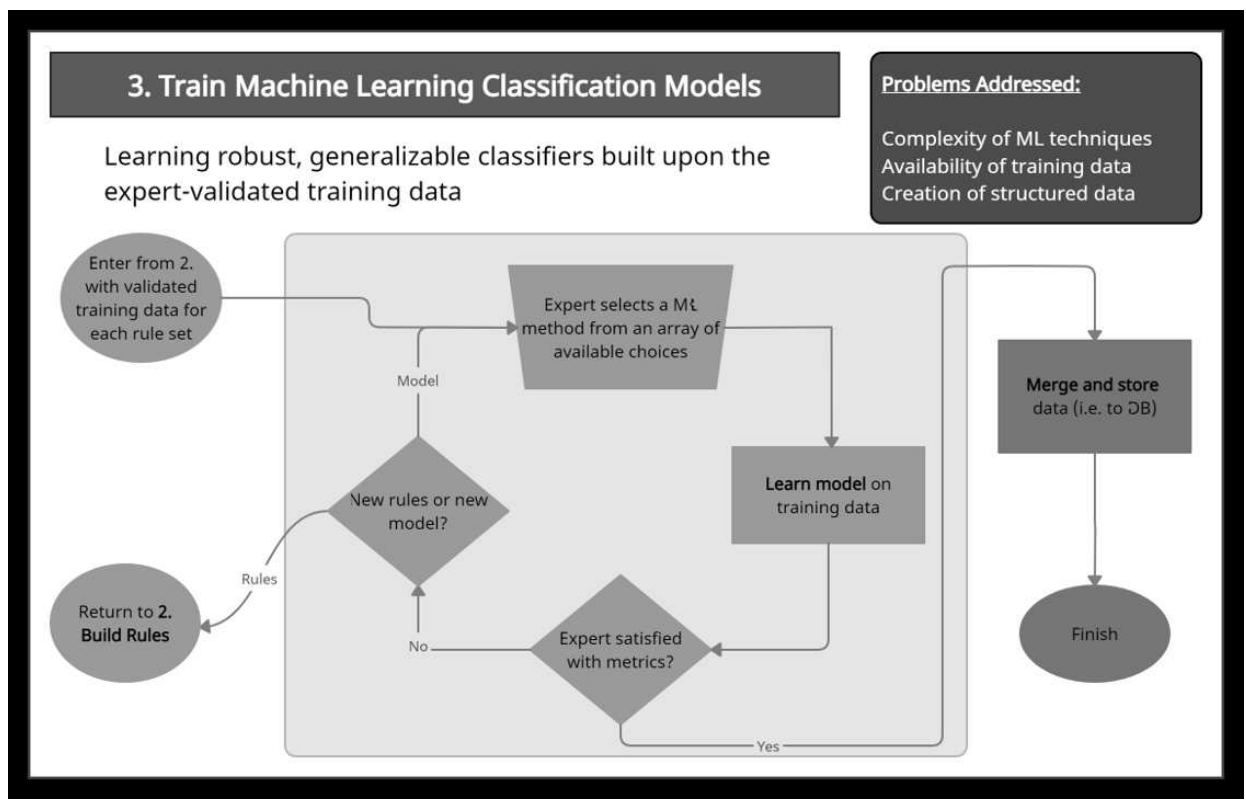}
    \caption{Third process in the CRAML Framework -- Train Classifiers}
    \label{fig:ml}
\end{figure}

Once the classifier(s) are trained, they can be deployed to classify the original data, i.e. the entire unstructured text corpus, or can be used on other sources of data. This represents the completion of the CRAML framework, as one can now build a structured, labeled dataset from the unstructured text of a document corpus using a ML model. This entire final process is illustrated in Figure \ref{fig:ml}.

\subsection{Evaluating the Model}

In this empirical context, we train a no poach classifier built with training data from  California. We use the classifier to build datasets for both California, in order to obtain a comparison between the rules-based approach and the ML approach using the same data. The ``no poach'' classifier is built with training data from a 10\% sample plus all of the no poach tag=1 observations in the full sample. In some cases, when the occurrence of positive or negative tags is very low, manual augmentations to training data such as this are required in order to receive acceptable model performance. We obtain an accuracy score of 0.99, precision of 0.97, recall of 0.96, and an overall F1-Score of 0.97. The harmonic mean of precision and recall (or F1-score) is based upon a comparison between the ML output and the training dataset. A common statistic in ML, with a maximum score of 1 when precision and recall are perfect, and 0 when either precision or recall is zero, the F1 score suggests a high level of accuracy and recall between the training dataset and results.

%We also apply the ML model to the Minnesota franchise document dataset, to analyze data outside of the main sample. 

\begin{figure}[ht]
    \centering
            \centering
            \includegraphics[width=\textwidth, scale=0.4]{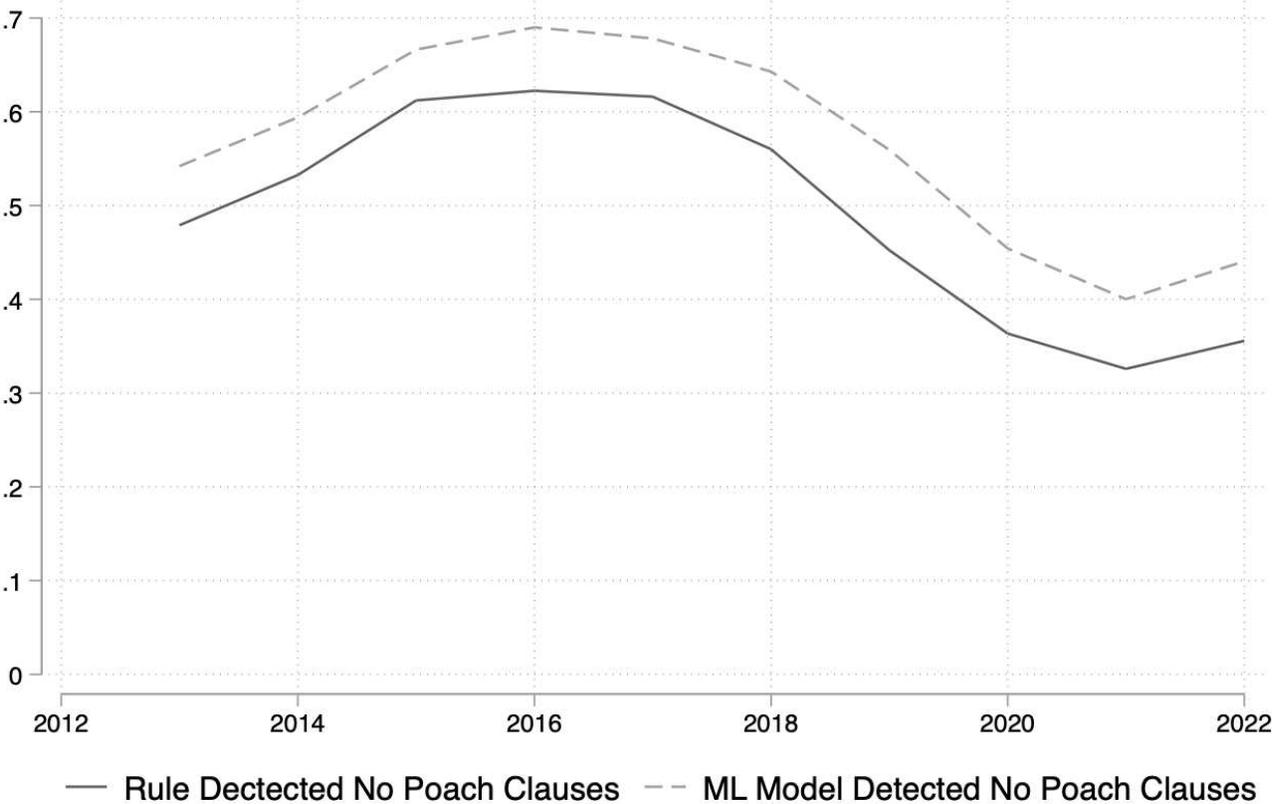}
   \caption{Analysis of Suspect No Poach Clauses in California Franchise Documents} 
   \label{fig:combofig}
\end{figure}

Figure \ref{fig:combofig} displays the percent of all rule-detected and ML-detected no poach clauses in the California Corpus within each year for each record filed from years 2013-2022 (with partial data for 2022). The pattern depicted with the California data is similar to the rule-based analysis. It shows that from 2015-2017 over 60\% of the records contained no poach clauses, after which there was a decline until the percent stabilizes below 40\%. This suggests that the interventions that followed \cite{krueger_theory_2022} had a large effect to decrease the prevalence of anti-competitive no poach language in franchise documents. The continued prevalence of these clauses, despite interventions from authorities, requires further investigation (in another paper): while the no poach clauses continue to exist, inspection of the text chunks reveals that while some remain the same as before the intervention of the Washington State Attorney General, while others now limit their applicability to more favorable jurisdictions and others restrict their application to highly compensated employees.

%Figure \ref{fig:combofigd}. depicts rule-detected and ML model trained classifier detected no poach clauses using the Minnesota corpus. While a lower share of FDDs in Minnesota contain no poach clauses, Minnesota illustrates the same post-2017  decline in no poach clauses. 

\paragraph{Practical Note on Model Accuracy} One noteworthy aspect of CRAML is that the above statistics reflect performance of the model at the ``chunk'' level. If an expert were looking to confirm that certain documents contain no poach clauses, the record is the level of analysis where they would want to begin their search. Given over 150,000 documents stored in nearly 13,000 records, this could be a daunting task. Moving to the level of the records, in California, the no poach classifier closely matches the rule-generated  results, with only 22 false negatives and 672 false positives in 12,922 records. Thus, at the record level, ML model achieves a 0.95 F1-score, 0.996 recall, 0.91 precision, and accuracy of 0.946. 

Further, an inspection of text chunks reveals that a significant portion of the false positives are in fact related to other language that accompanies no poach clauses and belong properly in the context of anti-competitive language in these documents: jurisdictional restrictions on no poach clauses, language that applies no poach clauses only to specific employees, and language that creates non-compete clauses. Precision at the level of the record would be 0.04 higher (0.95) if one considers these clauses to be relevant to a search for a broader construct of anti-competitive language. The results of CRAML are fully traceable to choices made in the construction of a rule set for the no poach classifier. As seen in Figure \ref{fig:combofig}, the ML classifier identifies more suspected no poach clauses then the rules, and this increases in 2021. Additional research using this corpus is  pinpointing fine-grained characteristics of language in these documents that restrict employee mobility. 

A final note: if one were dissatisfied with the model performance and sought one with higher recall or lower accuracy, or a more or less specific definition of the no poach construct, it would possible to ``steer learning'' and achieve a desired result by changing rule sets and thus re-shaping the training data. If desired, a user could add or remove all jurisdiction, narrow, and non-compete language from the no poach rule set in order to achieve a more discriminating classifier. A user could also combine all of the rules to make a single classifier that memorizes a larger pattern related to all types of anti-competitive language that appear in the documents.

%% file: discussion.tex
\section{Discussion}
\label{sec:discuss}
The CRAML framework makes several contributions. First, the framework enables the systematic and structured process of creating novel datasets from unstructured text corpora. Second, the result harmonizes advanced, automated information extraction and ML techniques with the input and expertise from manual analysis performed by supervising researchers. The exploration and iterative process performed during the context rule creation stage involves work that is not only difficult to perform automatically, but also that will arguably never come close to matching the diversity and expertise of human experts with domain knowledge. As a result, this intermediate stage in the framework incorporates a human element that is lacking in modern classification frameworks. 

%The potential for discovery and meaningful application of CRAML to unstructured text is demonstrated in several pilot projects completed while CRAML was being developed. Norlander and Erickson (in progress) discover the availability of remote work or work from home jobs in a corpus of over 290 million job advertisements. These ML classifiers could  widen the scope of opportunity for job seekers and employers with niche requirements/amenities. Norlander (in progress) describes the significance of no poach clauses that are plausibly illegal under both state and federal antitrust laws that ban efforts between firms to restrict competition. Because all the data for the no poach project is sourced from public documents, everything is replicable and transparent: from the raw corpus to the hand coded rules, the validation by independent third parties, and the ML model trained. 

\subsection{Future Directions}
\subsubsection{Technical Improvements}
Future research directions regarding the technical backbone of the CRAML framework should aim to bolster the extrapolation stage of the pipeline. Embeddings can be utilized to improve the transformation of context rules to datasets. With embeddings added to the extrapolation method, the need for training a ML classifier in the end may become obsolete.

A second area of future work involves building up tools to aid the domain expert in the initial, manual-driven stages of the CRAML process. Keyword extraction, topic modeling, automatic rule induction, and knowledge graphs could all be useful tools to support the domain expert express his or her worldview, as well as explore large text corpora in a richer way.

Although the focus is currently placed on ML techniques, the role of more advanced classifiers remains an open area of investigation. Deep Learning, in particular sequential models, could prove to be powerful in boosting the predictive capability of models trained on CRAML-generated datasets. Of course, this would come with the added overhead of extra training and parameter tuning, which must also be taken into account for future endeavors.

\subsubsection{Computational Social Science}
CRAML is designed with flexibility and web-based sources of text in mind.  A plug-in was created to integrate DocumentCloud (used by journalists and academics to post government documents) for the project related to no poach agreements. New plug-ins could draw data from web-based sources of text such as Twitter.  Niche ML classifiers built from user-data in a specific domain could address content moderation challenges at scale \cite{gillespie_content_2020,gorwa_algorithmic_2020}. To address a concern in \cite{porway_funding_2022}, government, civil society, and non-governmental organizations could develop context rule sets to capture text that is relevant to their interests and publish or license ML classifiers that provide useful feeds and classifications of information tracking on topics of societal concern. For academics, editors and reviewers could demand greater transparency when claims are made about data built form text. 

Concerns with proprietary data sources and privacy can be partially addressed by the information extraction methods used here. While additional efforts are needed, including the potential development of further tools for differential privacy with text, the extraction process is a step toward efforts to protect individuals against identification and protect the data owner from violation of proprietary information and trade secrets. For example, researchers have been barred from accessing records due to privacy and confidentiality concerns, but with careful selection of keywords, analysts could code chunks of extracted text and classify every document in a corpus without ever accessing the full or confidential information.\footnote{For example, researchers could use CRAML software to track the 20th Century use of residential restrictive covenants that barred Black, Jewish, Asian and other ethnic groups from living in certain neighborhoods. County commissioners of deeds have either barred researchers or severely limited access to digitized records due to concerns. For one project focused on Hennepin County, see https://mappingprejudice.umn.edu/.}

Finally, researchers in the humanities and social scientists can hopefully develop diverse and practical applications that uncover constructs and patterns currently ``hidden'' in unstructured text. The CRAML framework and software tools can be used to analyze text in any language. To the extent a construct is intra-subjectively reliable, it could be developed by a researcher into rules and scaled over vast amounts of unstructured text. To the extent that inter-subjective variance is high, we hope that by decreasing the time and effort needed to scale a framework, there will be many opportunities for diverse voices to produce contesting interpretative schemes that are transparent and replicable.

%\section{Conclusion}

% We contribute a method and software tools to transform unstructured text in a large document corpus to structured, labeled data ready for quantitative social science research and/or ML training. Context Rule Assisted Machine Learning (CRAML) is a structured text classification framework for building datasets from document corpora using a novel suite of software tools. We describe how to scale expert-generated schema over large volumes of text and produce labeled datasets that accurately characterize each document within a corpus. The process is a comprehensive approach to sampling, extracting, and analyzing text, and building context rules that produce structured, labeled datasets that can be used to annotate each document, or to train a Machine Learning (ML) classifier that learns a model to generalize beyond the confines of the context rules. We demonstrate CRAML’s efficacy at detection using a corpus of public sector contract documents, produce a novel training dataset and full dataset, and benchmark the results of context-rules based on precise string matching as well as trained ML classifiers.

%% file: appendixa.tex
\section{Appendix A: Using CRAML for a Literature Review}
\label{sec:appendixa}
Bibliographic research often begins with search for relevant papers using an academic search engine, but such search has a difficult time retrieving only the papers that use a specific empirical method or type of data, for example. To find papers that illustrate the challenges CRAML was created to address, we analyzed a corpus of all published papers from 5 top management journals from 2015-2021 (Academy of Management Journal, Administrative Science Quarterly,  Journal of International Business Studies, Organization Science, and Strategic Management Journal). We first assembled the corpus and PDFs in Zotero, and then exported the metadata. We converted all PDFs from these journals to text, minimally cleaned the text (lower case, removing punctuation, replaced abbreviations), and used CRAML software to find relevant papers using keywords, tags, and rules.

We deductively generated list of tags and keywords. The keywords.json file contians the tags and keywords used to extract all keyword-containing chunks of text. The context window was set to 6 words, yielding chunks of text of a maximum length of 13 words. After retrieving 13 word chunks, we examined n-grams in CRAML and developed rules files for each tag in order to classify the chunks that were relevant to our search. The following headings present the rules files used to classify which papers in the management literature contain relevant concepts - keywords alone generated numerous false positives. We note that this effort could be much further refined. Given the small size of the corpus and the variety of tags, machine learning was not desirable or attempted. CRAML was highly useful to iteratively narrowing a search for relevant literature, and we stopped when the results produced the relevant papers. As seen in the notes on the rule sets, keyword based approaches alone are insufficient to narrowing accurately, and researchers can benefit from a process of refinement that is context-aware and rule-based.

\subsection{``Is Text'' rules: Papers drawing from Text Corpora}

These rules that define the tag \emph{istext} or ``is text'' identify papers that refer to a text corpus or text data. In the first priority of rule-based classification, chunks containing the keywords ``text'', ``repository'', ``corpus'', and ``document'' are initially coded 0 (not \emph{istext}). The subsequent rule at priority level 1 classifies all chunks containing the exact matching text (e.g. ``text analyis'') 1 (is \emph{istext}). Given the \emph{istext} rules, the resulting metadata will report  the papers that discuss text data and corpora equal to 1 for \emph{istext}. The most relevant papers were found when combining the \emph{istext} tag with additional tags described below.

\begin{figure}[H]
    \centering
    \begin{tabular}{l|c|c}%
    \bfseries Rule & \bfseries Prio & \bfseries istext
    \csvreader[head to column names]{istext.csv}{}
    {\\\hline\rule\ & \prio & \istext}
    \end{tabular}
    \caption{\textit{istext} Rules}
\end{figure}

\subsection{ML rules}

To find papers that discuss supervised learning, we construct the simple \emph{ML} tag. To make Figure \ref{fig:frequency_of_words}, we focus only on observations where \emph{istext} and \emph{ML} are both equal to 1.

\begin{figure}[H]
    \centering
    \begin{tabular}{l|c|c}%
    \bfseries Rule & \bfseries Prio & \bfseries ML
    \csvreader[head to column names]{ml.csv}{}
    {\\\hline\rule\ & \prio & \ML}
    \end{tabular}
    \caption{\textit{ML} Rules}
\end{figure}

\subsection{NLP Rules}

We follow the same structure for the \emph{NLP} tag for finding papers that use NLP. For NLP, we also use regular expressions to find any chunk containing mention of the term ``embedding'' and one or more of ``word'', ``text'', or ``language.''

\begin{figure}[H]
    \centering
    \resizebox{\textwidth}{!}{
        \begin{tabular}{l|c|c}
        \bfseries Rule & \bfseries Prio & \bfseries NLP
        \csvreader[head to column names]{nlp.csv}{}
        {\\\hline\rule\ & \prio & \NLP}
        \end{tabular}
    }
    \caption{\textit{NLP} Rules}
\end{figure}

\subsection{Proprietary rules}

The keywords for proprietary data and models and the \emph{proprietary} tag extracted a large number of irrelevant chunks. To limit the focus to non-transparency in data and models, we use regular expressions to combine our base priority keywords with the added terms ``data'' or ``model'' to capture the papers that use or refer to proprietary data or models.

\begin{figure}[H]
    \centering
    \resizebox{\textwidth}{!}{
        \begin{tabular}{l|c|c}%
        \bfseries Rule & \bfseries Prio & \bfseries proprietary
        \csvreader[head to column names]{prop.csv}{}
        {\\\hline\rule\ & \prio & \proprietary}
        \end{tabular}
    }
    \caption{\textit{Proprietary} Rules}
\end{figure}

\subsection{Topic Modeling Rules}
For the topic modeling tag \emph{topic\_model}, we look for papers with chunks of text that refer to unsupervised learning, Latent Dirichlet Allocation (LDA), and topic models.

\begin{figure}[H]
    \centering
    \begin{tabular}{l|c|c}%
    \bfseries Rule & \bfseries Prio & \bfseries topic\_model
    \csvreader[head to column names]{tm.csv}{}
    {\\\hline\rule\ & \prio & \topicmodel}
    \end{tabular}
    \caption{\textit{Topic Model} Rules}
\end{figure}

\subsection{AI Rules}

For the tag \emph{AI}, we look for papers with chunks of text that refer to artificial intelligence. 

\begin{figure}[H]
    \centering
    \begin{tabular}{l|c|c}%
    \bfseries Rule & \bfseries Prio & \bfseries AI
    \csvreader[head to column names]{ai.csv}{}
    {\\\hline\rule\ & \prio & \AI}
    \end{tabular}
    \caption{\textit{AI} Rules}
\end{figure}

\subsection{The Results}

CRAML's extrapolate function yields a CSV file. Each row is a paper, and each column contains the metadata attributes of the paper (authors year, etc.) and each tag constructed in CRAML. For each paper, CRAML provides a 0/1 indicator of whether or not the paper contains a chunk corresponding to  the rules files above. Here, we present a table that aggregates the result of the CRAML analysis by year and theme:

\begin{table}
\begin{centering}
\begin{tabular}{| p{1.5cm} | p{2cm}  | p{2cm}  | p{1.5cm}  | p{1.5cm}  | p{2cm}  | p{1.5cm}  | p{1.5cm} |}
\hline
Year	& Proprietary Data or Models	 & Topic Models or LDA & 	ML & 	NLP	 & 	AI & 	CATA & Text or Document Corpus  \\ \hline
2015 & 19 & 2 & 0 & 1 & 0 & 53 & 7  \\
2016 & 11 & 0 & 1 & 0 & 0 & 68 & 10 \\
2017 & 18 & 0 & 0 & 0 & 2 & 69 & 9  \\
2018 & 20 & 2 & 0 & 2 & 2 & 73 & 17 \\
2019 & 13 & 3 & 2 & 1 & 3 & 82 & 18 \\
2020 & 17 & 2 & 2 & 1 & 1 & 67 & 18 \\
2021 & 19 & 6 & 7 & 3 & 2 & 80 & 31 \\
\hline
\end{tabular}
\end{centering}
\caption{Total Papers by Year that Reflect Rule Sets}
\end{table}

Because we are interested in particular in papers that use text corpora, to make the graph in Figure \ref{fig:frequency_of_words} and conduct our literature review, we filtered the metadata for only the papers that equal 1 for \emph{istext}. CRAML allowed us to find ``needles'' in a ``haystack'' of publications,   narrowing our search of the literature, allowing us to more closely examine the plausibly relevant papers, and read and incorporate relevant papers into our literature review. The data for Figure \ref{fig:frequency_of_words} that highlights the discussion of various text analysis techniques within these papers is presented below:

\begin{table}
\begin{centering}
\begin{tabular}{| p{1.5cm} | p{2cm}  | p{2cm}  | p{1.5cm}  | p{1.5cm}  | p{2cm}  | p{1.5cm}  | p{1.5cm} |}
\hline
Year	& Proprietary Data or Models	 & Topic Models or LDA & 	ML & 	NLP	 & 	AI & 	CATA & Text or Document Corpus  \\ \hline
2015 & 1 & 2 & 0 & 1 & 0 & 6  & 7  \\
2016 & 2 & 0 & 1 & 0 & 0 & 8  & 10 \\
2017 & 0 & 0 & 0 & 0 & 0 & 6  & 9  \\
2018 & 1 & 1 & 0 & 1 & 1 & 11 & 17 \\
2019 & 3 & 2 & 1 & 1 & 1 & 10 & 18 \\
2020 & 2 & 1 & 2 & 0 & 1 & 13 & 18 \\
2021 & 4 & 5 & 6 & 3 & 1 & 19 & 31 \\
\hline
\end{tabular}
\end{centering}
\caption{Subset of Papers Where \emph{istext}=1 by Year}
\end{table}

%% file: appendixb.tex
\section{Appendix B: Using CRAML to Build Classifiers from Job Advertisement Text}
\label{sec:appendixb}
One advantage of CRAML is that a classifier can be trained and validated using proprietary information but can be made available to other academic researchers for transparency, validation, and replication while preserving underlying proprietary information. To illustrate, we describe and publish nine classifiers from job advertisement text and outline our process for building and validating these. We describe a process for building open source research datasets drawn from proprietary text data with transparent rules and discuss a growing research ecosystem to develop useful classifiers drawn from the text of job advertisements.

% \paragraph{Accessing Chunks from the NLx Corpus}

% Initial chunks of text containing keywords being released are related to identifying barriers to employment in job advertisement text. Barriers to employment emerged as an area of need for study through discussions with non-profit labor market intelligence and workforce development organizations. Following extraction of the chunks, a good faith effort was made to remove any chunk that contains information that potentially ties a job ad to a specific firm, by using the Wasi and Flaeen dictionary of terms to identify and remove chunks that might contain a firm name (e.g., chunks with terms such as " co " or " inc "). This approach has serious effects on the underlying data in some cases (e.g., the guest worker visa-related tags use the keyword "visa" and many irrelevant chunks mention "visa inc").

% The keywords extracted, and the potential barriers, are listed below. These are released under a Creative Commons Attribution-NonCommercial-ShareAlike 4.0 International License. Users are asked to share a small amount of information about their intended use. Additional chunks and keywords may be extracted and released through an application and following a similar de-identification process.

\subsection{Building and Publishing New Classifiers from Job Ad Text}
\label{sec:jobad_tags}
Below, we describe the construction and performance of classifiers already constructed using job advertisement text data. The following variables are constructed from a random forest ML algorithm that classifies each job advertisement 0/1 on a particular variable. These variables are otherwise coded 0, but are equal to 1 if the job advertisement indicates:

\begin{itemize}
\item Work From Home. A job with the opportunity to work-from-home, including both full-time and hybrid work-from-home jobs.
\item Independent Contractor. A 1099 independent contractor job.
\item Professional License Required. A job requiring a professional license or formal apprenticeship.	
\item Union Presence. A job within a collective bargaining unit or a job that involves negotiating and working with unions (labor relations jobs, e.g.).
\item Citizenship Required. A job where the holder must be a U.S. citizen.
\item Work Authorization Required. A job where the job holder must be legally authorized to work in the U.S.
\item Visa Inclusivity. A job for which a firm is willing to sponsor a visa-holder for the position.
\item Visa Exclusivity. A job for which a firm is not willing to sponsor a visa-holder for the position.
\item EOE/AA Statement. A job ad that contains EOE/AA statement is present.
\end{itemize}

In building each classifier, close attention was paid to building reliable and valid measures. To construct each variable listed above, the 6 words before and after every appearance of a “keyword” (in column 2) of Table \ref{tab:appb1} were extracted from the text corpus of job advertisements.

\begin{table}
    \centering
    \begin{tabular}{|p{5cm} | p{12cm}|} 
    \hline \hline
    1. Variable	& 2. Keyword(s)	 \\
    \hline
    Work From Home &	``home", ``remote", ``telecommut", ``telework", ``virtual", ``videoconferenc", ``internet", ``anywhere" \\ \hline
    Independent Contractor	& “contract”	\\ \hline
    Professional License Required	& ``licens", ``credential", ``certifi", ``apprentice"  \\ \hline
    Union Presence	& ``collective", ``bargain", ``contract", ``union"  \\ \hline
    Government Contractor &	``affirmative", ``opportunity", ``eoe", ``contract"  \\ \hline
    Citizenship Required &	“citizen” \\ \hline
    Work Authorization Required	 & “authoriz”	\\ \hline
    Visa Inclusive &	“visa”, “1b”  \\ \hline
    Visa Exclusive &	“visa”, “1b”  \\ \hline \hline
    \end{tabular}
    \caption{Keywords for Initial Classifiers Built}
    \label{tab:appb1}
\end{table}

Research assistants and one lead researcher hand-coded a sample of the most frequently occurring 13-word ``text chunks'' 0 or 1 for each variable we wanted in the finished dataset. Based upon the hand-coded sample, context rules were initially drafted. Table \ref{tab:appb2} illustrates the hand-coding and manual validation process. For the work from home variable, independent hand-coding generated 1,524 increasingly lengthy context rules. Following extrapolation to a training dataset drawn from a random sample, RAs performed an additional round of independent hand-coding to arrive at a total of 2,901 hand-coded chunks for the work from home variable that are matched to the training dataset. Comparing known “true positives” via hand-coding and known “true negatives” from hand-coding to the training dataset, we are confident that the training dataset classifies at least 15,297 work from home chunks correctly. Because the process begins by feeding the training dataset all chunks containing the keywords, most chunks in the training dataset are equal to 0, and include many chunks containing irrelevant uses of the word ``home'': ``homework tutor'', ``home healthcare'', and ``homeland security'', e.g., all contain the keyword ``home'' after processing. This ``negative sampling'' of chunks is crucial for training ML models to distinguish between a 0 and 1. Because the focus is accuracy, or correctly identifying work from home jobs with a 1, it is possible that recall suffers, as this process misses isolated instances of work from home jobs with highly unusual text patterns.

%Context rules begin by calling all the keywords 0s at the base level of priority. Every chunk containing the word ``home'' is scored 0. The second priority level over-writes 0s by refining the keywords with additional context: “work from home” is scored 1. A third priority level overwrites a 0 for all text chunks containing the pattern “cannot work from home.” This elaboration of rules goes on until at the end, a dataset that was extrapolated based on context rules conforms precisely to the hand-coded chunks.

\begin{table}
    \centering
    \begin{tabular}{|p{5cm} | p{5cm} | p{5cm} |} 
    \hline \hline
    1. Variable		& 2. Hand-coded Chunks &	3. ML Sample N Matched with Hand-Coding \\ \hline
    Work From Home &	2,901 &	15,297\\ \hline
    Independent Contractor	& 1,287	& 8,771  \\ \hline
    Professional License Required &	11,107	& 15,665	 \\ \hline
    Union Presence	&	5,357 &	14,254  \\ \hline
    Government Contractor &	1,427 &	19,564  \\ \hline
    Citizenship Required &		418	& 242	 \\ \hline
    Work Authorization Required	& 890 & 1,888  \\ \hline
    Visa Inclusive &	6,796	& 3,660	 \\ \hline
    Visa Exclusive & 6,764 &	1,613 \\ \hline  \hline
    \end{tabular}
    \caption{Hand-coding and Matched ML Sample for Initial Classifiers Built}
    \label{tab:appb2}
\end{table}

Once a strategic sample of the training dataset provides a perfect match with the hand-coded data, we use the training dataset to train a ML model that predicts, e.g., if a given job is work-from-home. We tested logistic regression, na\"ive Bayes, and random forest models, and found that random forests perform best. Table \ref{tab:appb3} presents the size of a strategically selected sample of ML results used to assess accuracy. This sample size varies for each variable, as each variable occurs with different frequency in the raw data, and while initially drawn from a random sample of chunks containing the keywords, it is not random, but oversamples the most frequently occurring ``positive'' chunks.

Sample size needed to validate varies by the complexity of the rules needed to capture variation in the chunks that contain the keywords. For some variables, a large sample is necessary because of subtle variations in the text: there are many ways to indicate a ``work from home'' job, and as such, there are many keywords and many different types of chunks that contain the keyword. In complex cases, a domain expert must elaborate many rules to capture the variation in the text. For other variables, such as ``citizenship required,'' among chunks containing the keyword ``citizen,'' there is little variation in the text and only 63 context rules are required to handle the classification scheme envisioned. A smaller sample is acceptable in such cases because the frequency of key chunks in the sample of chunks are relatively large: correctly classifying the chunk ``citizenship is required'' 1 and almost everything else in the sample 0 captures most of the nuance. A very high proportion of the percentages in Column 2 of Table \ref{tab:appb3} not known to be true positives are true negatives.

The percent of known true positives in column 2 of Table \ref{tab:appb3} is based upon exact matches between hand-coded chunks and the results of a sample of ML tagged chunks. Finally, to assess the accuracy of the ML algorithm, column 3 reports the F-1 score. 

\begin{table}
    \centering
    \begin{tabular}{|p{5cm} | p{5cm} | p{5cm} |} 
     \hline \hline
    1. Variable	&	2. \% of ML Sample Confirmed True Positives &	3. F1 Score \\ \hline
    Work From Home &	92.1\% & 0.976 \\ \hline
    Independent Contractor	& 98.4\%	& 0.994 \\ \hline
    Professional License Required & 74.5\%	& 0.890 \\ \hline
    Union Presence	&	67.4\%	 & 0.874 \\ \hline
    Government Contractor &	39.6\%	& 0.996 \\ \hline
    Citizenship Required &		 69.4\%	& 0.976 \\ \hline
    Work Authorization Required	&	97\% &	0.994 \\ \hline
    Visa Inclusive 	& 80.6\%	& 0.890 \\ \hline
    Visa Exclusive 	& 76.0\% &	0.874 \\ \hline \hline
    \end{tabular}
    \caption{Percent of True Positives Confirmed by Hand-Matching and F-1 Scores}
    \label{tab:appb3}
\end{table}

While ``work from home'' is an omnibus measure of whether a job-holder can work from home, we earlier abandoned work on potential refinements such as hybrid work field jobs that permit work from home. Each step in this process has the potential for mistakes to slip into the classification process. While the ultimate validity rests on the opinion of other experts, the hand-coded files, the software, and large samples of text chunks can be published to encourage others' efforts. The data and software transparency will allow others to make their own models following their own process and could lead to better classifications. 

In the process described above, we paid additional attention to \textit{qualifiers}: words that have the ability to change the meaning of a text chunk in a significant manner. \textit{Keep} words are just that -- words that should be kept, even though they exist in a region to be trimmed away.

\subsubsection{Building Open Source Datasets} 

The framework adopted and the CRAML software tool facilitates a pipeline for  storing all (merged) classification output to a database of choice. In this particular application with the data mentioned here, a simple relational SQLite database with the following schema was used.

\begin{figure}[H]
\centering
\begin{tikzpicture}
    \node[align=center] (jobs) {
    %\nodepart{second}
    \underline{id} INTEGER PRIMARY KEY \\
    year INTEGER \\
    month INTEGER \\
    soc REAL\\
    onet REAL \\
    employer TEXT \\
    sector REAL \\
    fips TEXT \\
    title TEXT};
\end{tikzpicture}
\caption{Example Database Schema (for Jobs)}
\end{figure}
%\end{tabular}

Using this schema, a $jobs$ database was created where the novel, complete, merged, and classified dataset is stored. An example snippet of this database's contents is shown in Figure \ref{dbout}.

\begin{figure}[ht]
\centering
\begin{filecontents*}{db.csv}
id,vendor,effective-date,nopoach
app-7653,DHH Group Inc.,3/24/2017,1
app-15957,Apogee Learning LLC,9/20/2019,1
app-1853,The Agency Real Estate Franchising LLC,4/3/2015,1
app-7888,Winmark Corporation (Once Upon A Child),3/28/2017,1
app-18366,Atomium Inc.,6/25/2020,1
app-23189,Vision Source LLC,3/11/2022,0
app-22302,Brinker International Payroll Company L P (Chili's),11/1/2021,0
468744,Annex Brands Inc.,1/17/2013,0
app-14481,Gokhale Method Institute Inc.,4/11/2019,1
app-14643,Pizza Guys Franchises Inc.,7/22/2019,0
\end{filecontents*}
\csvautotabular{db.csv}
\caption{Random Sample from Jobs Database}
\label{dbout}
\end{figure}

Note that much of the data from the meta-data was left out for readability and for anonymization. The tag names have also been abbreviated, but they follow the same order as listed in Section \ref{sec:jobad_tags}. In essence, these tag columns represent the codification of the original text data into a structured dataset. This dataset, in turn, can be used for further research and analysis.

% Before we built the firm-level dataset of employment practices, we transform the firm name:
% \begin{itemize}
% \item We standardize all firm names using the \cite{wasi_record_2015}procedure and create “std\_name”. 
% \item We create a “parent\_name” variable that assigns all job advertisements in a subsidiary firm to the parent corporation using WRDS information on subsidiary corporate entities.
% \item For the S&P 1500, we perform additional hand-coding of variation in firm names when found in the NLx corpus. 
% \item For federal government agencies, we do additional hand-coding of naming conventions.
% \item For the S&P 1500, we hand-code and build crosswalks to GVKEY, CUSIP, TICKER.
% \end{itemize}

\subsection{Growing an Ecosystem for Labor Market Research from Text}

The pipeline outlined above regarding the ML models and rule sets we released can be scaled and other researchers can build their own text to data classifiers of job advertisements. NLx Research Hub data is proprietary and licensed to academic users for academic use, as well as state workforce agencies for research purposes. Access to the full corpus of job advertisements and the associated metadata requires an application and compliance with a (free) licensing agreement in order to protect sensitive information. However, releasing de-identified extracts of keyword-containing chunks of the text of job ads could be both fair use and encourage a community of academic researchers and citizens to develop ML classifiers.  Such text chunks, and not the full corpus, are all that is needed to build a niche classifier. The process of constructing a ML classifier could be done entirely in CRAML without ever needing access to sensitive information. 

Once built, if licensed for non-commercial academic use, scholars who obtain permission to access the underlying full text can use ML classifiers on the full text corpus and produce new open-source research datasets. Certain forms of data aggregation pose little risk: for example, occupation, state, commuting zone, and yearly summary data. With respect to firm-level data, aggregation is possible when exercising judgement to ensure no recent sensitive matters are publicly released (e.g. recent recruitment efforts that indicate adoption of a particular business strategy that might affect market valuation or competitors' strategy).

%% file: appendixc.tex
\section{Appendix C: The CRAML Pipeline} % Pipeline??
\label{sec:appendixc}
This appendix describes the CRAML methodology in detail with illustrations from the worked example introduced in Section \ref{sec:analysis}. 

\subsection{Extraction Algorithm}
In Algorithm \ref{ext}, we present the algorithm to extract context windows from a set of text documents, i.e. a corpus. This represents the important first step for handling large-scale unstructured text corpora by first extracting the \say{candidate diamonds} from the rough, in order to allow for the precise analysis of a domain expert.

\begin{algorithm}[htbp]
\caption{Extraction Algorithm}
\label{ext}
\begin{algorithmic}[1]
    \Require{\textbf{Text}: collection of text documents, \textbf{Keywords}: list of defined keywords}
    \Ensure{list of extracted files (location)}
    \State $all\_contexts$ = \{\}
    %\State $f$ = unzip($file$) \Comment{Unzip week file}
    %\State $xml$ = etree.iterparse($f$) \Comment{Parse xml tree}
    \For{$raw\_text$ in $documents$} \Comment{Note: done in parallel (unordered)}
            \State contexts = []
            \State $id$ = \textit{generate id}
            \If{any keyword in text}
                \State $cleaned$ = clean($text$) \Comment{text replace and regex cleaning}
                \For{$sentence$ in $cleaned$}
                    \If{any keyword in sentence}
                        \State $c$ = get\_context($sentence$) \Comment{get context window}
                        \State $contexts$.append($c$)
                    \EndIf
                \EndFor
            \EndIf
            \State $contexts$ = set($contexts$)
            \State $all\_contexts$[$id$] = $contexts$
    \EndFor
    \State $df$ = pd.DataFrame($all\_contexts$)
    \State $df$.to\_csv()
    \State \Return list of saved filenames
\end{algorithmic}
\end{algorithm}

\subsection{Keywords}
As displayed in Algorithm \ref{ext}, a list of defined keywords is required for extraction. The extraction algorithm will extract any chunk of text where one of the keywords appears, and store the chunk and metadata about its location in the corpus inside a CSV file for human analysis. Users must define a list of keywords for each \textit{tag}, which are given further elaboration in Section \ref{rules}.

Note that the extraction algorithm pulls all chunks where keywords are exact matched to a string of text. For example, the keyword \say{employ} will extract all of the variants that contain that exact keyword (employment, employer, employee, unemployment, etc.).

\subsection{The Output}
The result of running Algorithm \ref{ext} is a CSV formatted dataset containing chunks of text, with a minimum of two fields: id and text. Note that additional fields may also be included. The ID matches the identification number from the original data, and text represents a delimited list of \textit{context windows} around extracted keywords. In this case, a context window with a defined parameter $n$ refers the an extracted chunk of text with a maximum of $n$ words to the right and left of the keyword. If a punctuation mark is reached before $n$, then the context window will be shorter. Thus, with a $n$ of 12, context windows are a maximum of 25 word chunks. Context windows can be selected as either words or sentences; ie, a user can extract all keyword-containing sentences and the $n$ sentences to the left and right of a keyword. A sample from one extracted CSV file in Table \ref{tab:extract-example} illustrates this, with firm names changed: 

%\begin{figure}[htbp]
%\centering
%\csvautotabular{data/craml_ext_example.csv}
%\caption{Example extracted context windows}    
%\end{figure}

\begin{table}[htbp]
\centering
\begin{tabular}{|l|l|l|}
\hline
\textbf{id} & \textbf{firm} & \textbf{text} \\ \hline
10D4977B-000 & Acme Inc. 1 &  \{you shall not hire or permit ...\} \\
ADSPO15-0807 & Acme Inc. 2 &  \{will not hire an applicant...\} \\
CHR21009-CHS & Acme Inc. 3 &  \{a contractor shall not employ ....\} \\
99999-001 & Acme Inc. 4 & \{non solicitation agreements with ...\} \\ \hline
\end{tabular}%
\caption{Illustrative extracted context windows with fictional company names}
\label{tab:extract-example}
\end{table}

Note that for display purposes, only outputs relatively short in length were chosen. It is often the case that the extracted \say{text} is much longer in length, i.e. many contexts are extracted per contract. The output of the extract algorithm is an \say{intermediate} dataset of (ID, text) tuples, where the text is simply a long, '$\mid$' delimited string of all the extracted context windows. Looking at the example shown, one can see that in each extracted chunk of text, a keyword is present, along with its extracted context. 

\subsection{N-gram Exploration}
The first step towards building up rule sets is having all of the plausibly relevant text extracted. The exploration of n-gram structures within the extracted data ensures that the context window is correct, and if so, serves as a first step for construction of rules described in the following section. 

This is accomplished with the algorithm presented in Algorithm \ref{lr}. The result with a given $n$ is the a file with the enumerated occurrences of each relevant n-gram, sorted in descending order.

\begin{algorithm}[ht]
\caption{N-gram Exploration}
\label{lr}
\begin{algorithmic}[1]
    \Require{\textbf{Parent Directory}: where the extracted CSVs are located, \textbf{n}: desired n-gram size}
    \Ensure{file with enumerated n-grams}
    \State $global\_counts$ = \{\}
    \For{file in parent directory}
        \For{row $text$ in file}
            \State $split$ = $text$.split('$\mid$')
            \For{$chunk$ in $split$}
                \State $length$ = len($chunk$.split())
                \State $rules$ = $chunk$.split()[max(0, $length$/2 - n), min($length$, $length$/2 + n)]
                \State $global\_counts$.update(Counter($rules$))
            \EndFor
        \EndFor
    \EndFor
    \State \Return $global\_counts$.to\_csv()
\end{algorithmic}
\end{algorithm}

\subsection{Rule Creation: Building the Bridge}
\label{rules}
From the previous step, a researcher is now enabled to obtain a general picture of what context windows appear with the unstructured text data, as well as their relative occurrences. Using this, the researcher follows two decision processes: (1) the creation of a tag set, and subsequently (2) the creation of a rule set for each tag. Both sets are defined below.

\paragraph{Tags} A tag is a title given to a certain characteristic of a particular document. These tags can be easily and simply defined, and are binary in nature, i.e. 1 if true or 0 if not true. The collection of defined tags make up the \textit{tag set}. Each tag in the tag set is assigned \textbf{either} 1 \textbf{or} 0 for \textbf{every} unique document (i.e. unique ID) in the original data. Specifically for the franchise document example, the following tag is defined:

\begin{itemize}
    \label{tags}
    \setlength \itemsep{0em}
    \item \textit{nopoach} --  a non-solicitation clause that prohibits one franchise from recruiting workers from the another franchise. 
\end{itemize}

With this framework, one can define an arbitrary amount of tags; furthermore, exactly what these tags mean, ie. what definition they take on, is entirely up to the researcher. Binary classifications can be further broken down by defining additional \say{sub-tags}. As demonstrated below, an examination of the context surrounding non-solicitation suggests further tags for future research. At the data analysis stage, these can be combined to create new measurements of concepts involving multiple tags. Crucially, the meaning that is assumed by each tag is defined via its rule set, which is discussed next.

\begin{itemize}
    \label{tags_explained}
    \setlength \itemsep{0em}
     \item \textit{nopoach\_nohire} --  a no poach clause that includes a no hire clause.
     \item \textit{noncompete} --  an employee non-compete clause. 
     \item \textit{jurisdiction} -- language that limits the application of a no poach or non-compete to certain jurisdictions.
     \item \textit{narrow} -- language that limits the application of a no poach or non-compete to certain employees.
     \item \textit{no\_nopoach} -- language that limits the use of no poach or non-compete clauses - stating that these clauses are not enforceable against employees.
\end{itemize}

\paragraph{Rules} For each given tag, a researcher will now define a list of rules that encompasses the definition and characteristics of this tag. More concretely, a \textit{rule} is a chunk of any number of words. Therefore, a rule can be a single word, or even an entire sentence (up to the maximum length of the extracted chunks, here 25). A single rule set, defined within a corresponding file, can contain one or more tags. If more than one tag is contained within a rule set, this means that the tags that are bundled together are related enough such that some rules more overlap, or some rules may define one tag, while precluding the other. For each tag contained within a rule set, this tag is defined for each given rule, i.e. is assigned either 1 or 0. Finally, every rule in all rule sets must be assigned a \textit{priority} (\say{prio}), which indicates its relative order of operation compared to other rules within the same rule set. An example excerpt from a rule file is given in Table \ref{gc}.

\begin{table}[htbp]
\resizebox{\textwidth}{!}{%
\begin{tabular}{|l|l|l|}
\hline
\textbf{rule} & \textbf{prio} & \textbf{nopoach} \\ \hline
hire & 0 & 0 \\
shall not hire & 1 & 1 \\
will not hire & 1 & 1 \\
may not hire & 1 & 1 \\
you shall not hire or permit any third party or outside vendors to access or perform any service & 2 & 0 \\
may not hire an applicant who has a felony & 2 & 0 \\
will not hire any person regardless of medical marijuana card & 2 & 0 \\
shall not hire or promote anyone who may have contact with residents & 2 & 0 \\
not hire offer to hire or otherwise solicit any employee & 3 & 1 \\ \hline
\end{tabular}%
}
\caption{Excerpt from the rule set of the \textit{nopoach} tag in formative stages}
\label{gc}
\end{table}

Priority is important in the sense that it defines a hierarchy of how text chunks might be assigned certain tags. From the above example, one starts with every chunk containing \say{hire} to be \textit{nopoach}=0 at \textit{prio}=0. When, however, one encounters \say{may not hire} within a piece of text, this then is a plausible indicator that \textit{nopoach}=1. If a text chunk contains \say{may not hire} -- the subsequent  rule  \say{may not hire} assigned priority \textit{prio}=1 proceeds to classify it with \textit{nopoach}=1. Upon examination, that rule turns out to have exceptions. Moving on to a higher priority rule, \textit{prio}=2 can create exceptions to \textit{prio}=1 rules: for example, with \textit{prio}=2, a rule that states \say{may not hire an applicant who has a felony} overwrites the previous tag allocation, now assigning text chunks containing the string of text in the rule to \textit{nopoach}=0. In this way, priority is important to handling iterative coding work and possibly overlapping or contradicting rules.

The final important aspect of rule creation comes with the use of Regular Expressions, denoted by rules with the \say{REGEX:::} qualifier prepended to the regular expression itself. Instead of simple text matching for a rule, regex rules will instead check to verify if a particular text chunk satisfies the regular expression or not. Their usage allows for generalization, in situations where contexts are variable in length and content, yet possess the same general meaning.

Once the researcher feels that the rule set is saturated, it is important to ``prune'' rules and validate them through repeated extrapolation and adjustment. Context windows should be no longer than the longest rule, which should be no longer than necessary to precisely capture the tag. The final rule sets used in the franchise document analysis are available online.

\subsection{Extrapolation: Crossing the Bridge}
With this collection of rule sets (for all defined tags) in hand, a tool is now needed to translate the rules contained within to a workable dataset. Such a tool is useful because it takes the manually coded \textit{context rules} as input, and it then proceeds to \say{extrapolate} them into a training dataset from either a selected subset of extracted data files or the full dataset. Thus, we can \textit{cross the bridge} from expert-created rules to structured, annotated datasets.

\paragraph{The Subset} As mentioned, the extrapolation algorithm can operate on a selected subset of the entirety of the data, so as to create a representative training set, from which a robust, accurate classifier can be trained. This is accomplished by randomly and uniformly selecting documents from the original text files. In the case where a model will not be learned, i.e. where only the extrapolated dataset is desired, performing this sampling is not necessary. 

\paragraph{The Extrapolation Algorithm} The algorithm used to create a training data set for each tag is now defined. Note that this algorithm is run \textit{per} tag, meaning that the eventual output is a training set for each tag, which will then be used to train a classifier for each tag. The algorithm pseudocode is outlined in Algorithm \ref{extrapolate}.

\begin{algorithm}[ht!]
\caption{Extrapolation Algorithm}
\label{extrapolate}
\begin{algorithmic}[1]
    \Require{\textbf{Parent Directory}: where the subset CSV files are located, \textbf{Rules File}: a rule set for a given tag or tags, \textbf{s}: sampling rate, \textbf{do\_neg}: whether to perform negative sampling}
    \Ensure{training data set for given tag(s)}
    \State $results$ = []
    \For{file $f$ in subset}
        \State $data$ = read\_csv($f$).sample($s$) \Comment{sample desired fraction of data}
        \For{$text$ in data} \Comment{each '$\mid$' delimted extracted line}
            \For{rule $r$ in rule set}
                \State $match$ = []
                \State $plus$ = 0 \Comment{for negative sampling only}
                \For{x in $text$.split('$\mid$')}
                    \If{'REGEX' in $r$}
                        \If{regex.search($r$, $x$)}
                            \State $match$.append(($x$,1))
                            \State $plus$ += 1
                        \ElsIf{$plus$ $>$ 0 and $do\_neg$ == True}
                            \If{not any $ru$ in $x$ for $ru$ in rule set}
                                \State $match$.append(($x$,0))
                                \State $plus$ -= len(tags) \Comment{i.e. number of tags in rule set}
                            \EndIf
                        \EndIf
                    \Else
                        \If{$r$ in $x$}
                            \State $match$.append(($x$,1))
                            \State $plus$ += 1
                         \ElsIf{$plus$ $>$ 0 and $do\_neg$ == True}
                            \If{not any $ru$ in $x$ for $ru$ in rule set}
                                \State $match$.append(($x$,0))
                                \State $plus$ -= len(tags) 
                            \EndIf
                        \EndIf
                    \EndIf
                \EndFor
                \State random.shuffle($match$)
                \For{$m$ in match} \Comment{m[0] = text, m[1] = negative sample?}
                    \State $results$.append(text chunk, rule, priority-weighted length, tag encoding)
                \EndFor
            \EndFor    
        \EndFor
    \EndFor
    \State $training$ = DataFrame($results$)
    \State $training$ = $training$.sort(priority-weighted length).drop\_duplicates(chunk)
    \State \Return{$training$ as CSV}
\end{algorithmic}
\end{algorithm}

One important note about the extraction algorithm is the sampling rate. Since many files are included in the subset, running extrapolation on the entirety of this subset would result in quite large training datasets. To avoid this, only a random sampling of each subset file is taken, in order to ensure a manageable training dataset. Secondly, the concept of \textit{negative sampling} is incorporated into the algorithm. Such a concept allows for \say{negative}, non-keyword containing text chunks also to be included in the training data, in the case that a classifier that can discriminate between keyword- and non-keyword-containing instances is desired.

Ultimately, the main purpose of the extrapolation algorithm is to bridge the human-centric rule creation phase with the ensuing model training and classification stage. Concretely, the manually created rule sets that are the product of the former are converted to large, yet workable training data sets that are vital to the functioning of the latter. Thus, this transition from defined rule sets to classifier training is facilitated by such an extrapolation method.

As described above, a user may need to augment the training data produced from CRAML in order to improve model performance. Poorly performing models may reflect that the underlying tags are not well-defined, and need additional refinement, or simply that there are not enough positive or negative cases in the corpus to build a reliable classification model. 

\subsection{Baseline Testing Cycle: Boosting Real-World Applicability}

After an initial extrapolation, the expert engages in iterative testing to revise the rules to achieve a comprehensive and accurately coded dataset. The expert does this with output from the extrapolation. An excerpt (shortened for readability) of the \textit{nopoach} rules is provided in Table \ref{tab:training-data}.

%\begin{figure}[ht]
%\centering
%\csvautotabular{data/craml_training_data.csv}
%\caption{Some entries from the novel training data created for \textit{nopoach}}
%\end{figure}

\begin{table}[ht]
\resizebox{\textwidth}{!}{%
\begin{tabular}{|l|l|l|l|}
\hline
\textbf{id} & \textbf{chunk} & \textbf{rule} & \textbf{nopoach} \\ \hline
527557 & shall not recruit or hire any employee or former employee offranchisor or any & shall not recruit & 1 \\
781133 & of offenders page of monitoring center staff vendors shall not employ felons in & shall not employ felons & 0 \\
272571 & in item ahove you may not solicit customers from outside your territory without & may not solicit customers & 0 \\
487216 & or our designee you will not hire third party or outside vendors to & you will not hire third party or outside vendors & 0 \\
792020 & franchise lyou may not recruit or hire any employee or former employee of & may not recruit & 1 \\
714298 & non solicitation of employees employee agrees that during & non solicitation of employees & 1 \\ \hline
\end{tabular}%
}
\caption{Some entries from the novel training data created for \textit{nopoach} during testing}
\label{tab:training-data}
\end{table}

As can be seen, the extrapolated training data files contain the extracted text chunks, which rule \say{caught} the particular chunk, and finally the appropriate tag value. This file can be reviewed by the user to determine if the rules are working as intended. A sample of chunks that oversamples positive tags (=1) and provides a minimum number of cases per rule can also be sent at this point for blind review by third-party coders to validate and ensure inter-subject reliability. 

In the Table \ref{tab:training-data} example, early rules such as ``shall not recruit'', ``shall not employ'', and ``shall not hire'' were later abandoned due to the multiple exceptions to these rules as seen in the Table. Instead, discrete statements that clearly stated that an employee shall not be recruited or solicited or hired were tagged \textit{nopoach}=1

%For reference, Figure \ref{neg} shows an excerpt of training data when negative sampling is used (the \textit{rule} column is not relevant in the cases of a negative sample).

%\begin{figure}[ht]
%\centering
%\csvautotabular{data/traingcneg.csv}
%\caption{Training data for \textit{GovContract}, with negative sampling}
%\label{neg}
%\end{figure}

\subsection{Validation: Involving Third-Party Independent Coders}
Validation helps to ensure inter-rater reliability, accuracy, and precision. Involving third-parties at the stage at which rule sets are producing seemingly reliable results tests whether the tags are well-enough defined to be agreed to by third parties. The researcher gives an independent coder only a description of the desired tag, the text chunks, and deletes the rules and the CRAML-generated coding. The independent coder completes their review, and the researcher compares the results of the rule-set produced coding with the independent coder's coding. All discrepancies should be reconciled before proceeding further. 

Based upon the performance displayed by this last step in the process, or by an externally imposed requirement to repeat the process, the researcher can choose to revisit the tags and rule sets, once again performing an exploration to search for more representative rules. It is important to emphasize the necessity of judgement in this phase, as the decision to repeat this process is a subjective one. Incorporating hand-coding by independent research assistants can help greatly. Moreover, the encoding, i.e. tag values, given to each particular rule must come from knowledgeable, grounded reasoning, especially in cases where certain keywords serve different meanings in possibly very disparate contexts. Therefore, it is in this cycle where the most manual effort is required, but also where the crucial foundation to the rest of the framework is built.

\subsection{Model Training and Classification: Datasets in Action}
The final stage in the proposed framework is the classification itself, i.e. the mapping of a particular text chunk to its corresponding tag set encoding (0/1 for each tag). In order to accomplish this, classification models must be learned from the rule sets discussed in the previous section. This process is now outlined.

The presence of  context rules alone are not sufficient in the sense that they represent high precision classification rules that will always detect exactly what is described by the rules. The next step of using a classifier is to learn a model that not only detects these \say{simple} cases that can be caught by string matching or regular expressions, but rather one that can also learn in which \textit{general} contexts keywords appear which cause a certain tag to be true. This, therefore, motivates the need for comprehensive training data.

The final step in the testing cycle involves the training of a baseline classifier to test the performance of this classifier on an unseen test set. In this case, a simple Naive Bayes classifier was chosen. First, training instances are converted to TF-IDF vectors (introduced in more detail later), and then a Naive Bayes classifier is trained for each tag. Finally, output metrics are displayed, namely Accuracy, Precision, Recall, and F1. Using these metrics, a researcher can (roughly) evaluate the current performance of the classifier, which indicates the strength (\say{representativeness}) of the underlying training data, i.e. rules.

\subsubsection{Flexibility in Machine Learning Methods}
An important step towards the building of a general-purpose classification system was model selection. In particular, it becomes the task to identify the Machine Learning method best suited for the multi-label, binary classification task at hand. Crucial to note is this multi-label aspect, as it is certainly possible for a single document to have more than one tag attribute be present. As such, multiple candidates for classification models were chosen, all of which could handle this multi-label binary classification task. So far, only \say{shallow} Machine Learning models were tested.

The methods included:
\begin{itemize}
    \itemsep 0em
    \item \textbf{Naive Bayes} -- also used as the baseline classification method. Simple probabilistic classifiers using Bayes' theorem as a backbone. Relatively easy and efficient to train.
    \item \textbf{Logistic Regression} -- classification technique utilizing the logistic function (and a classification threshold) to predict the value of a dependent variable.
    \item \textbf{Stochastic Gradient Descent Classification} -- a misnomer in the sense that SGD does not actually perform the classification. Rather, SGD is used to optimize a linear model, in this case a Support Vector Machine.
    \item \textbf{Random Forest} -- a classifier using an ensemble of Decision Trees.
\end{itemize}

In a test of the various methods on a sample classification task (for the \textit{nopoach} tag), the classifiers performed as described in Table \ref{MLcomp}.

\begin{table}[ht]
\centering
\begin{tabular}{|l|c|c|c|c|}
\hline
 & \multicolumn{1}{l|}{\textbf{Acc.}} & \multicolumn{1}{l|}{\textbf{Precision}} & \multicolumn{1}{l|}{\textbf{Recall}} & \multicolumn{1}{l|}{\textbf{F1}} \\ \hline
\textbf{Naive Bayes} & 0.93 & 0.60 & 0.95 & 0.74 \\
\textbf{Logistic Regression} & 0.99 & 0.91 & 0.96 & 0.94 \\
\textbf{SGD SVM} & 0.98 & 0.91 & 0.92 & 0.92 \\
\textbf{Random Forest} & \textbf{0.99} & \textbf{0.97} & \textbf{0.96} & \textbf{0.97} \\ \hline
\end{tabular}
\caption{Performance of various ML classifiers}
\label{MLcomp}
\end{table}

In the case of large text corpora, particular in the example of no poach clauses in  state contracts, it may be the case that the positive cases, i.e. where a tag is assigned 1, are in the minority. As a result, the accuracy score becomes somewhat less important. More meaningful are the precision and recall scores. Essentially, precision measures the percentage of correctly identified positive cases, i.e. how many of the classified '1's are indeed truly positive cases. Recall measures that percentage of correctly identified positive cases among all positive cases in the true labels. Together, these two metrics can be succinctly summarized in the F1-score, which is the harmonic mean of the two.

As can be observed from Table \ref{MLcomp}, the Random Forest model shows superior performance with regards to all metrics. The same outcome was observed across the board with various rule sets, or rather their corresponding training data. As a result, the Random Forest was chosen to be the classification method of choice for the purposes of this dataset.

With the respect to the general-purpose nature of this text classification framework, it is important to emphasize that the choice of classification method here will not necessarily be the optimal choice for other applications. In addition to this, the utilization of more advanced and powerful methods could certainly prove to be beneficial, and this remains a topic for future research. In the end, though, the framework allows for essentially any model to be used, as long as it can be properly stored and loaded for use in the process pipeline. 

As part of the CRAML framework, the task of choosing a specific model is left to the user. This make sense due to the fact that different data, different domains, and different desired outcomes may require different models to be chosen. In the end, the focus of CRAML on the data \textit{creation} process allows for a flexibility in the choice of model to train on this data. Further details on this process can be found in the CRAML documentation.

\subsubsection{Outline of Model Training}
In order to adapt the use of classification models such as Random Forests for the classification of novel text datasets such as the one described throughout the previous sections, some considerations must be made. They are described below.

%\paragraph{TF-IDF} The first decision to be made is how to represent inherently unstructured text. Unfortunately, one cannot simply feed raw text into a classifier such as a Random Forest; rather, the text datasets must be numerically represented to allow for calculation and classification decisions thereupon. One of the simplest, and widely used, methods of doing this is the use of Term Frequency - Inverse Document Frequency vectors, TF-IDF in short. \textit{Term Frequency} takes into account the entire vocabulary (with size \textit{V}) of a given text dataset, and enumerates the occurrences of a particular word token within a 1 x \textit{V} vector. In order to add more meaningful information to these vector-valued representations, another score called \textit{Inverse Document Frequency} is incorporated. This score takes into account the relative uniqueness, or \say{rareness}, of a particular word token within one document (i.e. text chunk), compared to the universe of documents within a dataset. Inverse Document Frequency for a given token is calculated by dividing the total number of text entities in the dataset by the number of these entities than contain this particular token. In this way, this score acts almost as a weight which is applied to the Term Frequency. 

TF-IDF is utilized to convert a database of unstructured text entities into a matrix of numerically-valued vectors. Concretely, for a dataset containing $n$ \textit{documents} and a vocabulary of $m$ words, the resulting TF-IDF matrix is $n$ x $m$ in dimension. Through this vectorization of text, models can be trained and classification can now be performed.

As a point for future work, the utilization of more advanced (and potentially meaningful) text representations could lead to richer training data for the proposed CRAML framework.

\paragraph{Training - Grid Search}
With the training of ML models comes many tuneable parameters. In order to optimize the resulting classifier for each tag, a tuning stage was added into the framework, which determines the optimal parameters, e.g. for a Random Forest, given a particular dataset. 
%Thus, each Random Forest model could be tuned to its underlying dataset. The process for this was straightforward: (1) choose tuneable parameters, (2) perform broad random search, and (3) fine-tune via a grid search. For this specific application, the following parameters were chosen:
%\begin{itemize}
%    \setlength \itemsep{0em}
%    \item \textit{n\_estimators}: varies by tag
%    \item \textit{criterion}: \say{gini}
%    \item \textit{min\_samples\_leaf}: 2
%    \item \textit{min\_samples\_split}: 8
%\end{itemize}
Here it is important to emphasize the significance of this stage in the overall framework. As all potential incoming datasets may be different in nature, different parameter values may be needed to achieve the best performance possible. In order to best support the general-purpose nature of the proposed framework, performing fine-tuning of parameters is crucial.

\begin{figure}[ht!]
    \centering
    \fbox{\includegraphics[scale=0.9]{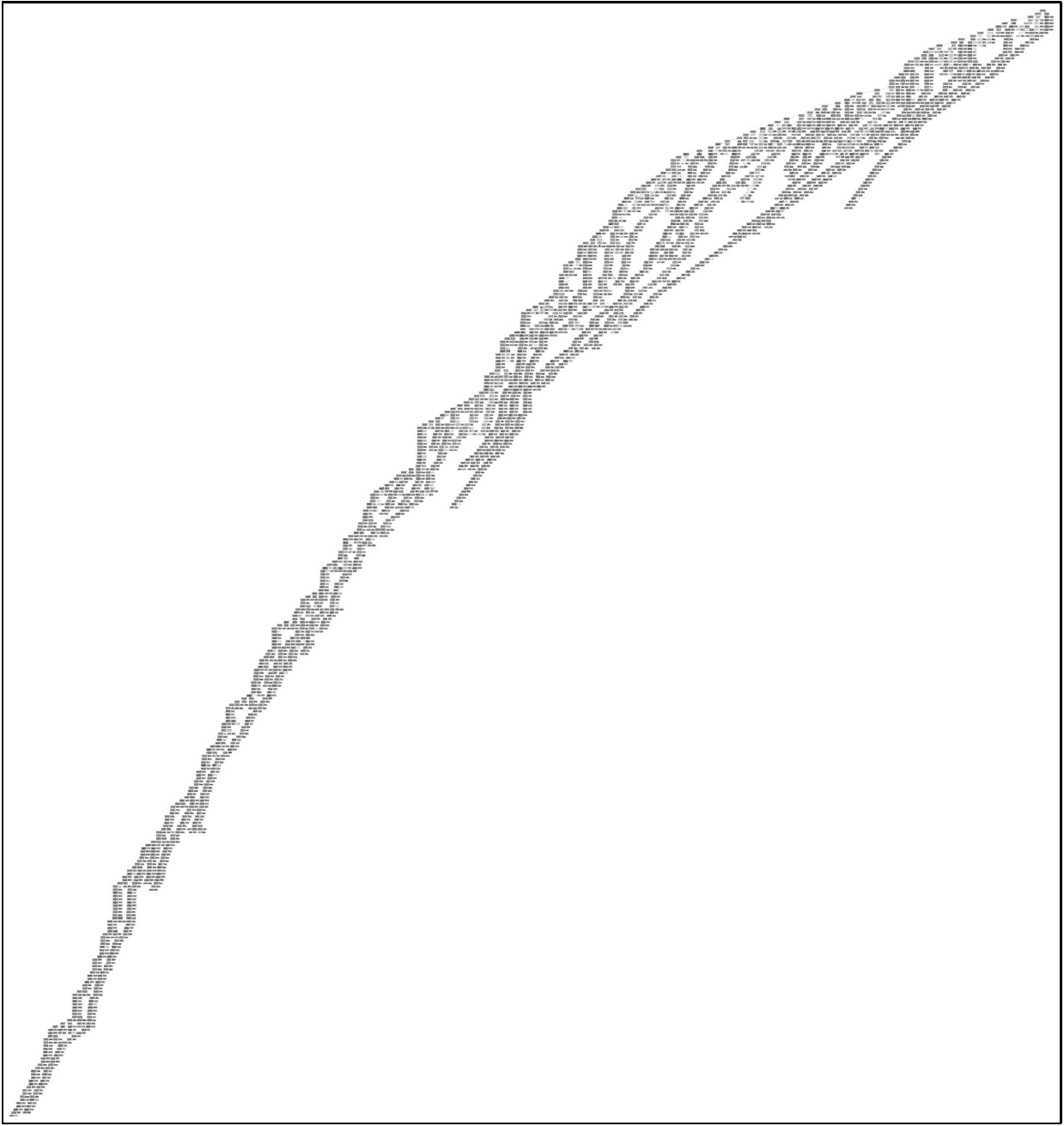}}
    \caption{One of the decisions trees within the \textit{nopoach} Random Forest}
    \label{gc_tree}
\end{figure}

\paragraph{Training - Purification}
As a final step to creating the classifiers for each tag, a purification process was run on the trained models, in order to reduce size, and as a result, classification time. This was accomplished via an available library, which eliminates unneeded dependencies and in the case of Random Forests, prunes the trees according to certain criteria. The effects of this process were quite dramatic, reducing most models to at least half the size. The potential complexity of Random Forests, and thus the need for the purification process, is visualized in Figure \ref{gc_tree}, which display a pre-purification decision tree.

\subsubsection{Classifier Integration}
Once a model for each tag is trained, it is then stored in pickle format, so that it can be later loaded and utilized within the framework. To do this, the user must manually enter the filename of each saved model into a JSON formatted file, which maps each tag to its corresponding classifier. 

%\begin{listing}%[float,language=json,firstnumber=1,caption={Classifier %Mapping},captionpos=b,label={clfmap}]
%\begin{minted}[frame=single,
%               framesep=1mm,
%               linenos=false,
%               xleftmargin=10pt,
%               xrightmargin=10pt,
%               tabsize=4]{js}
%{
%        "nopoach":{"clf":"RF_1.0-nopoach.pkl", "preproc":false},
%}
%\end{minted}
%\caption{Classifier Mapping}
%\label{clfmap}
%\end{listing}

Note how these tags match up with the mapping of tags to manually defined keywords. The naming convention used for the saved models was to specify the method used, the tag to be classified, the achieved F1-score, and the use of purification or not.

A significant decision that was made along the design process of the framework was whether to utilize the negative sampling scheme that was left as an option in the extrapolation process (Algorithm \ref{extrapolate}). This decision would most directly affect the function of the classifiers. More concretely, without negative sampling, the classifier for each tag would be to \say{classify a tag as 0 or 1, provided that the chunk to be classified contains a tag-defined keyword}. On the other hand, performing the the negative sampling process introduces random, non-keyword-containing text chunks into the model training process, thereby learning these instances into the given model. Here, the classification task becomes \say{classify any given piece of text as 0 or 1, regardless of its content}. With these two approaches in mind, it was ultimately decided to implement the former, thereby only advancing a text chunk to the classification stage if it indeed contained a keyword. If not, the classification was automatically made to be 0. It is quite important to stress that this design decision was made knowing the makeup of the textual data, as well as the characteristics of the tags to be classified. In the end, in order for a tag to be assigned a 1, a necessary condition is for it to contain a tag-specific keyword. Otherwise, it is not possible to obtain the classification of 1. The classification process, therefore, is in place to \say{weed out} keyword-containing text chunks that actually do not lead to the desired tag attribute. In other domains, i.e. with other text datasets created via this framework, this line of reasoning may not hold absolutely, and an argument could be made for the use of negative sampling. In this way, it is once again paramount for (a team of) researchers to define their tags thoroughly and explore the makeup of the underlying data.

One final component is the option of a preprocessing step for each classifier. The motivation is that in the text chunks, which are relatively large in length compared to the single keyword within, there might exist much information, i.e. words, that do not play a role in a particular tag at hand. Particularly with tags in which the pertinent information sits close to the keyword, any other irrelevant information will only serve to confuse the learning of a classification model. With this in mind, the proposed preprocessing step will serve to \say{trim} the left and right ends of a given text chunk, thus reducing the amount of text noise within this chunk. In this process, though, two considerations must be made. Firstly, if certain qualifier words, such as negations, exists within the regions to be trimmed, this could potentially be vital information lost. Secondly, there may be specific known words or phrases that appear that usually appear in the proximity of a keyword, yet might also be trimmed away. For these reasons, two more mappings, following the structure of previous ones where an entry exists for each tag, are defined. \textit{Qualifiers} involves the words that have the ability to change the meaning of a text chunk in a signficant manner. \textit{Keep} words are just that -- words that should be kept, even though they exist in a region to be trimmed away. %An example qualifier mapping for the job posting data is provided in Listing \ref{qualmap}:

%\begin{listing}%[float,language=json,firstnumber=1,caption={Qualifier %Mapping},captionpos=b,label={qualmap}]
%\begin{minted}[frame=single,
%               framesep=1mm,
%               linenos=false,
%               xleftmargin=10pt,
%               xrightmargin=10pt,
%               tabsize=4]{js}
%{
%        "nopoach": ["no", ...]
%}
%\end{minted}
%\caption{Qualifier Mapping}
%\label{qualmap}
%\end{listing}

With these definitions and the preprocessing steps, the pseudocode is outlined in Algorithm \ref{preproc}.

\begin{algorithm}[ht]
\caption{Preprocessing Algorithm}
\label{preproc}
\begin{algorithmic}[1]
    \Require{\textbf{Training CSV}: file that contains the training data to be trimmed, \textbf{Tags}: tags to be processed, \textbf{Qualifiers}: for each tag, \textbf{Keeps}: for each tag, \textbf{TRIM}: target context size}
    \Ensure{trimmed training data}
    \State $new\_data$ = []
    \For{$text$ in training data}
        \State $index$ = get\_index($text$) \Comment{get index of keyword within chunk}
        \For{$q$ in Qualifiers}
            \State $found$ = search\_for($q$, $text$) \Comment{find qualifiers in chunk}
            \If{$found$}
                \State $text$ = qual\_prepend($found$, $text$) \Comment{prepend all found qualifiers to words in chunk}
            \EndIf
        \EndFor
        \State $lower$ = max(0, $index$ - TRIM)
        \State $upper$ = min(len($text$), $index$ + TRIM)
        \State $new\_text$ = $text$.split()[$lower$:$upper$]
        \State $keep$ = []
        \For{$k$ in Keeps}
            \If{$k$ in $text$}
                $keep$.append($k$)
            \EndIf
        \EndFor
        \State $new\_text$ = $keep$ + $new\_text$
        \State $new\_data$.append($new\_text$)
    \EndFor
    \State \Return $new\_data$
\end{algorithmic}
\end{algorithm}

In the testing with this preprocessing stage, the $TRIM$ parameter was chosen to be 2, resulting in chunks of length 5, not including possibly kept words. In the end, this stage was excluded from the framework in the context of the job posting data, for with many tags there was actually an observed decline in classification performance. This sheds light on the importance of relatively faraway context words from the main keyword for many of the tags. Again, this may differ with other datasets, so testing with this preprocessing stage is recommended.

\subsection{Storing}
The final process in the framework is to consolidate the classification results, merging them with the original data to be stored for later retrieval and analysis. %Note that the merging step is specific to the case of the job posting data, as well as any future applications where information, i.e. the \say{meta-data} from the original data is desired in the final output.

%\subsubsection{Merge} As mentioned, in the case of this application-specific data, a merging process needs to take place in order to consolidate two datasets. The output of the classification stage represents the classified job posting entries from all inputted weekly data files. In Section \ref{complete}, the monthly meta-data was introduced, and it was decided that this decided should also be included in the final output data. Because both the classification output and the complete data are indexed by the same unique job posting ID scheme, a merge of these two datasets is straightforward. In particular, a left join of the classified weekly data on the monthly complete data is performed. In this way, job postings not included in the (extracted) weekly data, i.e. job postings containing no keywords, are also included, by with encoding vectors of only 0s.

The framework facilitates for the storing of all classification outputs to a database of choice. In this particular application with the data mentioned here, a simple relational SQLite database with the schema in Figure \ref{nopoach_schema} was used.

\begin{figure}[H]
\centering
%\begin{tikzpicture}
%\node[basic] (nopoach) {nopoach
%\nodepart{second}
\underline{id} TEXT PRIMARY KEY \\
vendor TEXT \\
effective-date DATE \\
type TEXT \\
text TEXT
%};
%\end{tikzpicture}
\caption{Example Database Schema (for \textit{nopoach})}
\label{nopoach_schema}
\end{figure}

Using this schema, a $nopoach$ database was created where the novel, complete, and classified dataset could be stored. An example excerpt of this database's contents is shown in Figure \ref{dbout2}.

Note that some of the meta-data, including the text, was left out for readability. In addition, we mask the firm name here in order to avoid drawing attention to the specific detail as opposed to the results of the process. In essence, the tag column represents the codification of the original text data into a structured dataset. This dataset, in turn, can be used for further research and analysis.

\begin{figure}[ht]
\centering
\begin{filecontents*}{db2.csv}
id,vendor,effective-date,nopoach
app-7653,Acme Inc. 1,3/24/17,1
app-15957,Acme Inc. 2,9/20/19,1
app-1853,Acme Inc. 3,4/3/15,1
app-7888,Acme Inc. 4,3/28/17,1
app-18366,Acme Inc. 5,6/25/20,1
app-23189,Acme Inc. 6,3/11/22,0
app-22302,Acme Inc. 7,11/1/21,0
468744,Acme Inc. 8,1/17/13,0
app-14481,Acme Inc. 9,4/11/19,1
app-14643,Acme Inc. 10,7/22/19,0
\end{filecontents*}
\csvautotabular{db2.csv}
\caption{Random Sample from the \textit{nopoach} Database}
\label{dbout2}
\end{figure}